\documentclass{article}

\usepackage{microtype}
\usepackage{graphicx}
\usepackage{subfigure}
\usepackage{booktabs} 

\usepackage{paralist}

\usepackage{hyperref}
\usepackage{amsmath}
\usepackage{amssymb}
\usepackage{mathtools}
\usepackage{amsthm}
\usepackage{multicol}
\usepackage{multirow}
\usepackage{bm}
\usepackage{color}
\usepackage{mathabx}

\usepackage[accepted]{icml2024}

\usepackage{amsmath}
\usepackage{amssymb}
\usepackage{mathtools}
\usepackage{amsthm}

\usepackage[capitalize,noabbrev]{cleveref}

\theoremstyle{plain}

\theoremstyle{definition}

\theoremstyle{remark}

\newcommand{\cameraR}[1]{\textcolor{black}{#1}}

\usepackage[textsize=tiny]{todonotes}

\icmltitlerunning{Amortized Equation Discovery in Hybrid Dynamical Systems}

\begin{document}

\twocolumn[
\icmltitle{Amortized Equation Discovery in Hybrid Dynamical Systems}



\icmlsetsymbol{equal}{*}

\begin{icmlauthorlist}
\icmlauthor{Yongtuo Liu}{uva}
\icmlauthor{Sara Magliacane}{uva}
\icmlauthor{Miltiadis Kofinas}{uva}
\icmlauthor{Efstratios Gavves}{uva}
\end{icmlauthorlist}

\icmlaffiliation{uva}{University of Amsterdam}
\icmlcorrespondingauthor{Yongtuo Liu}{y.liu6@uva.nl}

\icmlkeywords{Machine Learning, ICML}

\vskip 0.3in
]



\printAffiliationsAndNotice{}

\begin{abstract}  
Hybrid dynamical systems are prevalent in science and engineering to express complex systems with continuous and discrete states.
To learn laws of systems, all previous methods for equation discovery in hybrid systems follow a two-stage paradigm, i.e. they first group time series into small cluster fragments and then discover equations in each fragment separately through methods in non-hybrid systems.
Although effective, these methods do not take fully advantage of the commonalities in the shared dynamics of multiple fragments that are driven by the same equations.
Besides, the two-stage paradigm breaks the interdependence between categorizing and representing dynamics that jointly form hybrid systems. 
In this paper, we reformulate the problem and propose an end-to-end learning framework, i.e. Amortized Equation Discovery (AMORE), 
to jointly categorize modes and discover equations characterizing the dynamics of each mode by all segments of the mode.
Experiments on four hybrid and six non-hybrid systems show that our method outperforms previous methods on equation discovery, segmentation, and forecasting.
\end{abstract}

\section{Introduction}
Complex systems in science and engineering 
often exhibit behaviors and patterns that change over time.
Hybrid dynamical systems~\cite{van2007introduction} characterize these systems by continuous time series which are interleaved with structural changes producing discrete modes.
For instance, consider the motions of antelopes in a herd and how these suddenly change in the presence of lions.
Hybrid systems are researched widely with applications in epidemiology~\citep{keeling2001seasonally}, legged locomotion~\citep{holmes2006dynamics}, robotics~\citep{cortes2008discontinuous}, the designs of cyber-physical systems~\citep{sanfelice2016analysis}, and systems with interacting objects~\citep{liu2023graph}. 

A major challenge with hybrid dynamical systems is that one cannot know a priori the number of possible modes and when the switching happens within them.
The dynamic modes might alternate from one to another constantly and at any time, due to either internal mechanisms or external influences.
When modeling generalized time series as hybrid dynamical systems, it is thus crucial that we categorize the complex dynamics into the most likely discrete modes while characterizing the continuous motion dynamics in between.

\begin{figure*}[t]
 \centering
 \setlength{\tabcolsep}{1pt}
 \includegraphics[width=.85\linewidth]{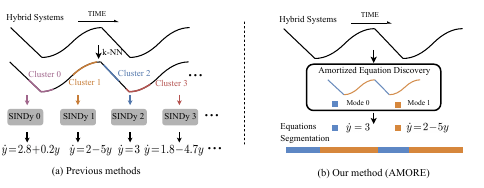}
 \vspace{-12pt}
 \caption{(a) Previous methods for equation discovery in hybrid dynamical systems typically follow a two-stage paradigm, i.e. they first group time series into small cluster fragments and then apply methods proposed in non-hybrid systems, e.g. SINDy~\citep{brunton2016discovering} to discover equations in each fragment separately. (b) Different from all previous methods, we reformulate the problem and propose a one-stage end-to-end learning framework, Amortized Equation Discovery (a.k.a. AMORE), to jointly categorize hybrid systems into discrete modes and discover equations characterizing motion dynamics of each mode based on all segments belonging to the mode.}
 \label{teaser}
 \end{figure*}

Another challenge with characterizing dynamics in hybrid systems, especially physical ones, is that predictive models are often not interpretable.
We are often interested in the underlying laws that govern the dynamics, thus preferring analytic models, usually in the form of closed-form 
ordinary
differential equations.
Equation discovery from first principles is a challenging problem in all fields of science.
To bypass expensive and cumbersome targeted experimentation, researchers have explored using data-driven methods for equation discovery of systems from observations~\cite{langley1981data,lemos2023rediscovering}. They
distill parsimonious
equations from data and find that compared with black-box neural networks, learned equations can provide insight into the essential dynamics of systems and tend to generalize better~\cite{lutter2019deep,karniadakis2021physics}.


Equation discovery for hybrid dynamical systems
has been a topic of interest for a long time 
~\cite{vidal2003algebraic,ozay2008sparsification,bako2011identification,ohlsson2013identification}. 
Recently, \citet{mangan2019model,novelli2022boosting}
proposed methods for equation discovery in non-linear hybrid systems. Both methods consist of two stages: they first group time series fragments into a large number of small cluster fragments and then apply an equation discovery 
method proposed in non-hybrid systems,
e.g. SINDy~\citep{brunton2016discovering}, to discover equations in each fragment separately.
The separate multi-stage processing limits the potential performance because it does not leverage the commonalities across fragments from the same mode and splits learning into two separate stages,
i.e. categorizing and then representing motion dynamics which jointly form hybrid systems.

In this paper, we reformulate the problem of equation discovery in hybrid dynamical systems and propose a one-stage end-to-end learning framework, Amortized Equation Discovery (a.k.a. AMORE), to jointly categorize motion dynamics and discover equations by modeling categorical modes and mode-switching behaviors within systems.
Equations are discovered to characterize the dynamics of each mode based on all segments that are assigned to the mode, by learning combinations of candidate basis functions and encouraging parsimony.
To model switching behaviors, inspired by REDSDS~\cite{ansari2021deep}, we infer latent categorical variables, i.e. mode variables, to categorize motion dynamics into discrete modes and learn probabilistic transition behaviors within them.
Equations, mode variables, and mode-switching behaviors are jointly learned in the proposed end-to-end learning framework by maximizing the system observation likelihood. 
We also consider another challenge in previous methods for equation discovery for hybrid systems: they are limited to single-object scenarios where the dynamics of only one object or objects as a whole are considered.
We extend our method to multi-object scenarios, AMORE-MIO, where multiple objects interact with each other and change their dynamics accordingly.
Extensive experiments on single- and multi-object hybrid systems demonstrate the superior performance of our method on equation discovery, segmentation, and forecasting. The code and datasets are available at \href{https://github.com/yongtuoliu/Amortized-Equation-Discovery-in-Hybrid-Dynamical-Systems}{https://github.com/yongtuoliu/Amortized-Equation-Discovery-in-Hybrid-Dynamical-Systems}.



\section{Related Work}
\paragraph{Equation discovery in hybrid dynamical systems.}
Prior works focus on the simplest hybrid dynamical models, i.e. piece-wise affine systems with linear transition rules~\cite{vidal2003algebraic,ferrari2003clustering,roll2004identification,juloski2005bayesian,paoletti2007identification,ozay2008sparsification,bako2011identification,ohlsson2013identification}. Recently, ~\citet{mangan2019model,novelli2022boosting} relieve these constraints and propose methods for non-linear hybrid systems. Among them, Hybrid-SINDy~\citep{mangan2019model} proposes a two-stage method, i.e. it first uses k-NN to group time series into small cluster fragments and then discovers governing equations separately in each fragment by models proposed in non-hybrid systems, e.g. SINDy~\citep{brunton2016discovering}. 
Based on Hybrid-SINDy, \citet{novelli2022boosting} also follows\ a two-stage paradigm while introducing the number of discontinuities in hybrid systems as a known prior for better performance.
Although effective, these two-stage methods learn the mode of each segment individually and do not leverage the commonalities across segments.
In this paper, we reformulate the problem and propose an amortized end-to-end learning framework to jointly categorize modes, discover equations, and learn mode-switching behaviors.

\vspace{-6pt}
\paragraph{Equation discovery in non-hybrid dynamical systems.}
Many methods have been proposed for equation discovery in non-hybrid dynamical systems. \citet{bongard2007automated} and \citet{schmidt2009distilling} leverage genetic programming~\citep{koza1994genetic} to discover nonlinear differential equations from data.  SINDy~\citep{brunton2016discovering} uses symbolic sparse regression on a library of candidate model terms to select the fewest terms required to describe the observed dynamics. Several methods extend this approach to new settings, e.g. identifying partial differential equations~\citep{rudy2017data}, considering additional physical constraints~\citep{loiseau2018constrained}, including control signals~\citep{kaiser2018sparse}, and introducing integral terms for denoising~\citep{schaeffer2017sparse}.
These methods cannot be directly applied to hybrid systems because they cannot model an unknown number of modes and unknown mode-switching behaviors.

\vspace{-6pt}
\paragraph{Switching dynamical systems.} 
Switching dynamical systems refer to the same systems as hybrid dynamical systems, but highlight different aspects in the literature. Hybrid systems denote systems with a mixture of continuous and discrete states, while switching dynamical systems highlight the switching behaviors of system states.
Many methods focus on switching linear dynamical systems where they set matrix calculations to model linear state transitions~\cite{ackerson1970state,ghahramani2000variational,oh2005variational}. Recently, switching non-linear dynamical systems model state transitions as neural networks, e.g. SNLDS~\citep{dong2020collapsed}, REDSDS~\citep{ansari2021deep}, and GRASS~\citep{liu2023graph}. While effective in modeling state-switching behaviors, they cannot discover closed-form equations from data. To categorize dynamics, our method is inspired by previous switching dynamical systems~\cite{dong2020collapsed,ansari2021deep,liu2023graph} to infer latent mode variables. Differently, our method can jointly discover parsimonious closed-form equations to characterize dynamics and infer the values of the mode variables.

\section{Equation Discovery in Dynamical Systems}

\label{sec: Equations for one motion mode}

In dynamical systems, the 
dynamics can be expressed by sets of differential equations in the form:
\begin{align}
\dot{\mathbf{y}_t} \coloneqq \frac{d\mathbf{y}_t}{dt} = \mathbf{f}(\mathbf{y}_t).
\label{eq: differential equation}
\end{align}
Equation discovery in dynamical systems is the task of learning the function $\mathbf{f}: \mathbb{R}^M\rightarrow \mathbb{R}^M$ from time-series observations $\mathbf{y}=\{\mathbf{y}_1, \cdots, \mathbf{y}_T\}$ where each state $\mathbf{y}_t=[y_t^1, \cdots, y_t^M] \in \mathbb{R}^M$. Following SINDy~\citep{brunton2016discovering}, 
we approximate each dimension $\dot{y}_t^m$ for $m \in [M]$ of $\dot{\mathbf{y}}_t$ in Eq.~\eqref{eq: differential equation} as
\begin{align}
\dot{y}_t^m = \frac{dy_t^m}{dt} = f_m(\mathbf{y}_t) \approx \Theta(\mathbf{y}_t)\xi_m,
\label{eq: approximate differential equation}
\end{align}
where $\Theta(\mathbf{y}_t) = [\theta_1(\mathbf{y}_t), \theta_2(\mathbf{y}_t), \cdots, \theta_P(\mathbf{y}_t))]$ is a set of candidate basis functions and  $\xi_m$ is a sparse vector indicating which of these function terms are active in characterizing the dynamics. 
We encourage the sparsity of $\xi_m$ based on Occam's razor principle, where the simplest model is likely the correct one.
Ideally, we could encourage this principle by minimizing the $L_0$ norm of the coefficients and solving the following constrained minimization problem
\begin{equation}
\mathop{\rm{min}}\limits_{\xi} ||\xi||_{0} \;\;\; {\rm subject \; to} \;\; || \Theta(\mathbf{y}_t)\xi - \dot{\mathbf{y}}_t || \leq \epsilon,
\label{eq: L0-norm sparsity-promoting optimization}
\end{equation}
where $\epsilon$ is a hyperparameter representing maximal optimization errors. 
The $L_0$ regularization 
penalizes the number of non-zero entries to encourage sparsity. However, optimization under this penalty is computationally intractable due to the non-differentiability and the combinatorial nature of all possible states. Various continuous relaxation methods are proposed in the literature to handle the optimization problems of $L_0$ norm, e.g. $L_1$, $L_2$, etc. As our focus in this paper is not to design advanced optimization methods, we utilize the simple and effective $L_1$ norm to optimize Eq.~\eqref{eq: L0-norm sparsity-promoting optimization}.

We implement the coefficients $\xi_m$ as learnable weights in neural networks.
We set the polynomial degree as $D$ and use a set of learnable weights $\mathbf{w}= [{w}_{1},\cdots,{w}_{C}]$ to model the coefficients of $C$ candidate basis functions. For instance, if the observation $\mathbf{y}_t = [a, b]$ is a two-dimensional vector and we set the polynomial degree $D$ as 2, the candidate basis polynomial functions are $\Theta(\mathbf{y}_t) = [1, a, b, a^2, b^2, ab]$. In this case, $C=6$ and $\mathbf{w}= [{w}_{1},\cdots,{w}_{6}]$.

\section{Equation Discovery in Hybrid Systems}
\label{sec: Equation Discovery in Hybrid Systems}
Hybrid dynamical systems produce generalized time series with continuous states and discrete events that need to be modeled, featuring multiple modes that represent different types of dynamics.
Instead of learning a single equation for each dimension $m \in [M]$, as described in Sec.~\ref{sec: Equations for one motion mode},
we learn $K$ sets of equations for each dimension $m$ that represent $K$ different modes in hybrid systems. We first introduce how we model mode-switching behaviors and then introduce our whole framework for equation discovery in hybrid systems. 

\vspace{-6pt}
\paragraph{Mode variables.} To model modes and mode-switching behaviors in hybrid systems, inspired by REDSDS~\citep{ansari2021deep}, we introduce latent categorical variables, i.e. mode variables, to learn categorical distributions of modes and index each set of equations representing each type of dynamics. 
Specifically, mode variables are discrete variables $\mathbf{z}\coloneqq \mathbf{z}_{1:T}$, where each $z_t \in \{1, \dots, K\}$ categorizes the mode at time step $ t \in \{1, \dots, T\}$.

\vspace{-6pt}
\paragraph{Count variables.} Besides mode variables, we also model latent count variables to learn the duration distributions of each mode. Count variables can help us avoid frequent mode switching, thanks to the fact that mode durations typically follow a geometric distribution~\citep{ansari2021deep}.
They are modeled as discrete categorical variables $\mathbf{c}\coloneqq \mathbf{c}_{1:T}$, where each state $c_t \in \{1, \dots, d_{\rm max}\}$ explicitly models the run-length of the currently active mode at time $t$ and $\rm d_{max}$ is the maximal number of steps before a mode switch. 
Count variables $c_t$ are incremented by 1 when the mode $z_t = k$ stays the same at the next time step $z_{t+1} = k$, or they reset to 1 when mode $z_t = k$ switches to another one $z_t \neq k$. 


\vspace{-6pt}
\paragraph{Mode-specific equation discovery.} 
Each mode $k \in [K]$ is assigned its own set of candidate basis functions $\Theta_k(\mathbf{y}_t)$ and learnable coefficient weights $\mathbf{w}_k$, which we will use to discover its equation.
For instance, at time $t$, the mode variable $z_t = k$ indexes the candidate basis function $\Theta_k(\mathbf{y}_t)$ and the learnable weights $\mathbf{w}_k$, which together define the equation representing the dynamics of mode $k$. In practice, different modes share the same candidate basis functions, i.e. $\Theta_j(\mathbf{y}_t)=\Theta_{k}(\mathbf{y}_t), \forall j,k \in \{1,\cdots, K\}$, unless we have some prior knowledge of the hybrid system. However, the learnable coefficient weights of different modes are individual and never shared, i.e. $\mathbf{w}_j \neq \mathbf{w}_{k}, \forall j,k \in \{1,\cdots, K\}$.
We collect all the candidate basis functions in a single vector $\Theta(\mathbf{y}_t) = (\Theta_1(\mathbf{y}_t), \cdots, \Theta_K(\mathbf{y}_t))$ and similarly we collect all learnable coefficient weights $\mathbf{w} = (\mathbf{w}_1, \cdots, \mathbf{w}_K)$.

\begin{figure}[t]
 \centering
 \setlength{\tabcolsep}{1pt}
 \includegraphics[width=.90\linewidth]{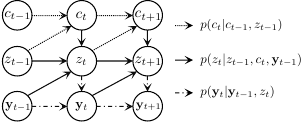}
 \caption{Generative model for amortized equation discovery. 
 $p(c_{t}|c_{t-1},z_{t-1})$ and $p(z_t|z_{t-1}, c_{t}, \mathbf{y}_{t-1})$ are count and mode transition probabilities, respectively. 
 $p(\mathbf{y}_t|\mathbf{y}_{t-1},z_t)$ denotes the observation transition probability where equations are discovered to characterize the  dynamics of each mode.
 }
 \label{AMORE}
 \end{figure}

 \begin{figure*}[t]
 \centering
 \setlength{\tabcolsep}{1pt}
 \includegraphics[width=.92\linewidth]{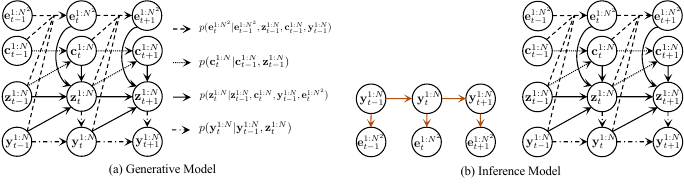}
 \vspace{-8pt}
 \caption{(a) Generative model of AMORE-MIO. 
 $p(\mathbf{e}_{t}^{1:N^2}\!|\mathbf{e}_{t-1}^{1:N^2}\!\!,\mathbf{z}_{t-1}^{1:N},\mathbf{c}_{t-1}^{1:N},\mathbf{y}_{t-1}^{1:N})$, $p(\mathbf{c}_{t}^{1:N}\!|\mathbf{c}_{t-1}^{1:N},\mathbf{z}_{t-1}^{1:N})$,
 $p(\mathbf{z}_{t}^{1:N}|\mathbf{z}_{t-1}^{1:N},\mathbf{c}_{t}^{1:N}\!\!,\mathbf{y}_{t-1}^{1:N},\mathbf{e}_{t}^{1:N^2})$, and 
 $p(\mathbf{y}_{t}^{1:N}|\mathbf{y}_{t-1}^{1:N},\mathbf{z}_{t}^{1:N})$ denotes the edge, count, mode, and observation transition probabilities, respectively. Equations are modeled at $p(\mathbf{y}_{t}^{1:N}|\mathbf{y}_{t-1}^{1:N},\mathbf{z}_{t}^{1:N})$ which characterize object-shared and mode-specific  dynamics.
 (b) Inference model of AMORE-MIO. Left: posterior approximate inference of edge variables $\mathbf{e}_{t}^{1:N^2}$. Right: Exact inference of discrete mode and count variables $\mathbf{z}_t^{1:N}$ and $\mathbf{c}_t^{1:N}$ based on observations $\mathbf{y}_t^{1:N}$ and the approximate edge variables $\mathbf{e}_{t}^{1:N^2}$. Orange arrows 
 denote the approximate inference flow.}
 \label{AMORE-MIO}
 \end{figure*}

\vspace{-5pt}
\paragraph{Generative model for AMORE.} Assuming Markovian dynamics, the joint generative probability of hybrid systems in our model is described as
\begin{align}
p(\mathbf{y}, \mathbf{z}, \mathbf{c}) = 
\underbrace{p(\mathbf{y}_1|z_1)
\,p(z_1)}_{\rm{Initial\,\,States}} \cdot 
 \prod_{t=2}^{T}\bigg[ 
 p(\mathbf{y}_t|\mathbf{y}_{t-1},z_t) \notag \\
p(z_t|z_{t-1}, c_{t}, \mathbf{y}_{t-1})
p(c_{t}|c_{t-1},z_{t-1})\bigg]
\label{eq:joint-full-single-object}
\end{align}
In the initial states, count variables are ignored as they are always 1 when starting. $p(z_1)$ is the initial distribution over all possible  modes. $p(\mathbf{y}_1|z_1)$ models the initial observation states conditioned on the initial modes. For later time steps $t \geq 2$, the count transition probability $p(c_{t}|c_{t-1},z_{t-1})$ models how the count variables at time $t$ change over time depending on their previous values and mode variables at time $t-1$. The mode transition probability $p(z_t|z_{t-1}, c_{t}, \mathbf{y}_{t-1})$ models mode-switching behaviors on how modes switch at time $t$ conditioned on the previous mode and observation states at time $t-1$ as well as the updated count state $c_{t}$ at time $t$. 
The observation transition probability $p(\mathbf{y}_t|\mathbf{y}_{t-1},z_t)$ models how the observations at time $t$ are influenced by their previous values at time $t-1$ conditioned on the updated mode variables at time $t$. Equations are amortized and learned at $p(\mathbf{y}_t|\mathbf{y}_{t-1},z_t)$ by all segments of each mode to characterize mode dynamics.
More specifically, conditioned on motion mode $z_t = k$, $p(\mathbf{y}_t|\mathbf{y}_{t-1},z_t)$ first indexes a set of candidate basis functions $\Theta_k$ and coefficient weights $\mathbf{w}_k$, which are used together to obtain derivatives $\dot{\mathbf{y}}_{t-1}=\Theta_k \cdot \mathbf{w}_k$ of $\mathbf{y}_{t-1}$ at time $t-1$. With known time intervals $\Delta_t$, we finally achieve $\mathbf{y}_{t}=\dot{\mathbf{y}}_{t-1} \cdot \Delta_t + \mathbf{y}_{t-1}$ assuming the  dynamics do not change much in short time intervals. For inference of the latent mode and count variables, we conduct exact inference similar to the forward-backward algorithm in HMM~\citep{eddy1996hidden}. The graphical model for the exact inference is the same as the generative model, which is illustrated in Figure~\ref{AMORE}.
The neural network implementations and details of the inference model are in Appendix~\ref{app: Neural Network Implementation of AMORE} and~\ref{app: Inference Model of AMORE}.


Learnable parameters of AMORE are optimized by maximizing the observation likelihood with sparse regularization on coefficient weights $\mathbf{w}$ of candidate basis functions
\begin{align}
\mathcal{L}_\textrm{AMORE} &= \rm -log\,p_{\theta}(\mathbf{y}) + ||\mathbf{w}||_{1} \notag \\
&= -\mathbb{E}_{p_\theta(\mathbf{z},\mathbf{c}|\mathbf{y})}\left[ {\rm log}\,p_\theta(\mathbf{y},\mathbf{z},\mathbf{c})\right] + ||\mathbf{w}||_{1}.
\end{align}
The derivatives of the training objective and further expansions over time are detailed in Appendix~\ref{app: Derivation of Training Objective of AMORE}.

\section{Equation Discovery in Multi-object Hybrid Systems}


While equation discovery in hybrid dynamical systems has been researched in single-object scenarios, the more general setting of systems with multiple potentially interacting objects 
is an unexplored yet natural setting.
In this section, we elaborate on how our model can be extended for multi-object scenarios, and present AMORtized Equation discovery in MultI-Object hybrid systems, a.k.a. AMORE-MIO. We first introduce how our method models interactions and then introduce the whole framework of AMORE-MIO.


\vspace{-5pt}
\paragraph{Edge variables.} Assume that $N$ objects and $K$ motion modes exist in multi-object hybrid systems. Inspired by~\citet{kipf2018neural,liu2023graph}, we model interactions between objects by latent edge variables $\mathbf{e}=\mathbf{e}_{1:T}^{1:N^2}=\{e_t^1, \cdots, e_t^{N^2}\}_{t=1}^T$ including self-loop, thus totally $N^2$ for $N$ objects.
For each pair of objects, interactions $e_t^{m\rightarrow n}$ model whether object $m$ interacts with object $n$ at time $t$.
The edge variables are modeled in a latent temporal graph $\mathcal{G}_t=(\mathcal{V}_t, \mathcal{E}_t)$, where edges $e_t^{m\rightarrow n} \in \mathcal{E}_t$ and nodes $\mathcal{V}_t$ summarize states of objects. For instance, $\mathbf{v}^m_t=\{z^m_t, c^m_t, \mathbf{y}^m_t\}$ for $\mathbf{v}^m_t \in \mathcal{V}_t$ defines one graph node summarizing states of object $m$ at time $t$ which includes observation $\{\mathbf{y}^m_t\}$ and latent states $\{z^m_t, c^m_t\}$.
Edge $e_t^{m\rightarrow n}$ signals interaction relationships between node $\mathbf{v}_t^m$ and node $\mathbf{v}_t^n$ in graph $\mathcal{G}_t$. 

\vspace{-5pt}
\paragraph{Object-shared and mode-specific equation discovery.} We set the number of all possible motion modes as $K$ in multi-object hybrid dynamical systems. The $K$ motion modes are shared across $N$ objects, which are implemented by $K$ sets of candidate basis functions $\Theta(\mathbf{y}_t) = (\Theta_1(\mathbf{y}_t), \cdots, \Theta_K(\mathbf{y}_t))$ and learnable coefficient weights $\mathbf{w} = (\mathbf{w}_1, \cdots, \mathbf{w}_K)$. Each mode $k \in [K]$ has its own $\Theta_k(\mathbf{y}_t)$ and $\mathbf{w}_k$ as in Sec.~\ref{sec: Equation Discovery in Hybrid Systems}.
\cameraR{Thus there are K sets of learnable weights for learning dynamics of K modes across N objects. Both the time and space complexity of AMORE-MIO regarding learnable weights of basis functions is $\mathcal{O}(K)$.}
For instance, the mode variable $z_t^n=k$ of the object $n$ at time $t$ indexes $\Theta_k(\mathbf{y}_t)$ and $\mathbf{w}_k$ which together form equations to represent the dynamics of mode $k$.
Different from single-object scenarios, the mode-switching behaviors of each object are not only influenced by their own evolving nature but also by external influences of potentially interacting objects. 
We model the influences of interactions on the mode-switching behaviors between objects, which are detailed in the following generative model of AMORE-MIO.

\paragraph{Generative model for amortized equation discovery in multi-object settings.} Assuming Markovian dynamics, the joint generative probability of multi-object hybrid systems in AMORE-MIO is calculated as
\begin{align}
&p(\mathbf{y},\mathbf{e},\mathbf{z},\mathbf{c})=\underbrace{p(\mathbf{y}_{1}^{1:N}|\mathbf{z}_{1}^{1:N})p(\mathbf{z}_{1}^{1:N})}_{\rm{Initial\,\,States}}\cdot \notag \\
&\prod_{t=2}^{T} \!\bigg[p(\mathbf{y}_{t}^{1:N}|\mathbf{y}_{t-1}^{1:N},\mathbf{z}_{t}^{1:N})p(\mathbf{z}_{t}^{1:N}|\mathbf{z}_{t-1}^{1:N},\mathbf{y}_{t-1}^{1:N},\mathbf{c}_{t}^{1:N}\!,\mathbf{e}_{t}^{1:N^2})\!\! \notag\\ 
    & p(\mathbf{c}_{t}^{1:N}|\mathbf{c}_{t-1}^{1:N},\mathbf{z}_{t-1}^{1:N})p(\mathbf{e}_{t}^{1:N^2}|\mathbf{e}_{t-1}^{1:N^2},\mathbf{c}_{t-1}^{1:N},\mathbf{z}_{t-1}^{1:N},\mathbf{y}_{t-1}^{1:N})\!\bigg]
\label{eq:joint-full-multiple-object}
\end{align}
where the initial states and count transition probability are defined as in Eq.~\ref{eq:joint-full-single-object} of single-object scenarios but with $n$ objects. For later time steps $t\geq 2$, the edge transition probability $p(\mathbf{e}_{t}^{1:N^2}|\mathbf{e}_{t-1}^{1:N^2},\mathbf{c}_{t-1}^{1:N},\mathbf{z}_{t-1}^{1:N},\mathbf{y}_{t-1}^{1:N})$ models how the edge variables evolve depending on all the states at the previous time step. We model the influences of interactions on the mode transition probability $p(\mathbf{z}_{t}^{1:N}|\mathbf{z}_{t-1}^{1:N},\mathbf{y}_{t-1}^{1:N},\mathbf{c}_{t}^{1:N},\mathbf{e}_{t}^{1:N^2})$, which characterizes the mode-switching behaviors of multi-object hybrid dynamical systems. Based on the updated modes of each object, the observation transition probability $p(\mathbf{y}_{t}^{1:N}|\mathbf{y}_{t-1}^{1:N},\mathbf{z}_{t}^{1:N})$ can be factorized over objects $\prod_{n=1}^{N}p(\mathbf{y}_{t}^{n}|\mathbf{y}_{t-1}^{n},z_{t}^{n})$ where equations of each mode are amortized and learned by all segments from all objects belonging to the same mode.
The further expansion over objects of the joint generative probability is in Appendix~\ref{app: Expansion of Generative Model over Objects of AMORE-MIO}.
For inference of latent mode, count, and edge variables, we conduct posterior approximate inference for edge variables $q_{\phi_e}(\mathbf{e}|\mathbf{y})$ conditioned on observations $\mathbf{y}$, and then conduct exact inference of mode and count variables $p_\theta(\mathbf{z},\mathbf{c}|\mathbf{y},\tilde{\mathbf{e}})$ conditioned on observations $\mathbf{y}$ and the approximate edge variables $\tilde{\mathbf{e}} \sim q_{\phi_e}(\mathbf{e}|\mathbf{y})$.
The generative and inference models of AMORE-MIO are illustrated in Fig.~\ref{AMORE-MIO}.
Learnable parameters of AMORE-MIO are optimized by maximizing the evidence lower bound with sparse regularization on the learnable coefficient weights
\begin{align}
&\mathcal{L}_\textrm{ AMORE-MIO} = -{\rm log}\,p_\theta(\mathbf{y}) + \notag \\
&\;\;\;\;\;\;\;\;D_{K\!L}\left[q_\phi(\mathbf{z},\mathbf{c},\mathbf{e}|\mathbf{y})\,\|\,p_\theta(\mathbf{z},\mathbf{c},\mathbf{e}|\mathbf{y})\right] + ||\mathbf{w}||_{1}
\end{align}
Neural network implementations,  the derivations and the detailed inference model are in Appendix~\ref{app: Neural Network Implementation of AMORE-MIO}, ~\ref{app: Derivation of Optimization Objective of AMORE-MIO}. and~\ref{app: Inference Model of AMORE-MIO}.
\section{Experiments}

We extensively validate our method on 10 dynamical systems.
Specifically, we validate on single-object scenarios using the Mass-spring Hopper dataset, and the Susceptible, Infected and Recovered (SIR) disease dataset from Hybrid-SINDy~\citep{mangan2019model}.
We validate on multi-object scenarios using the ODE-driven particle dataset and Salsa-dancing dataset from GRASS~\citep{liu2023graph}. 
Further, we test the robustness of our methods on non-hybrid systems using datasets of the Coupled linear, Cubic oscillator, Lorenz' 63, Hopf bifurcation, Seklov glycolysis, and Duffing oscillator from~\citet{course2023state}.
Detailed settings of the datasets are in Appendix~\ref{app: datasets}.

\begin{figure*}[t]
 \centering
 \setlength{\tabcolsep}{1pt}
 \includegraphics[width=.95\linewidth]{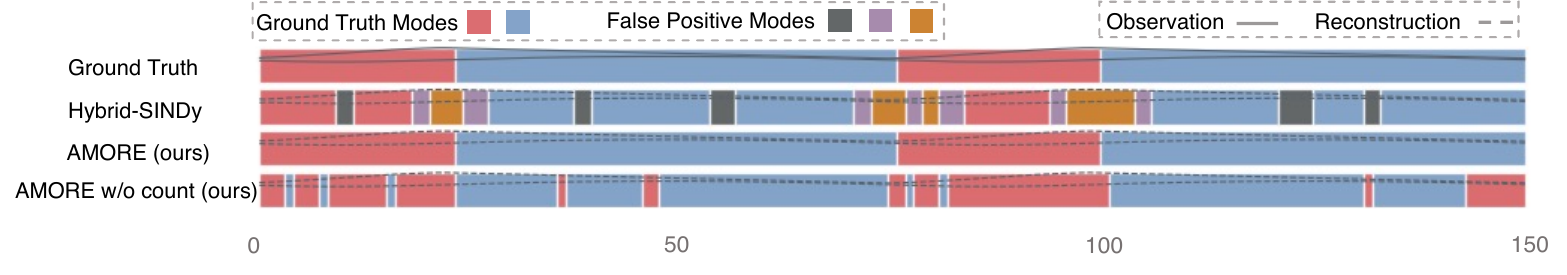}
 \vspace{-12pt}
 \caption{Qualitative time series segmentation results of AMORE compared to Hybrid-SINDy~\citep{brunton2016discovering} on the Mass-spring Hopper dataset. For Hybrid-SINDy, we aggregate the discovered equations with the same number of coefficients as one mode. We can see that with joint learning of modes and equations, AMORE can categorize the exact number of modes and achieve superior segmentation results with fewer switching errors.}
 \label{Qualitative time series segmentation results of AMORE compared to Hybrid-SINDy}
 \end{figure*}

 \begin{table}[t]
	\centering
        \footnotesize
	\tabcolsep=0.100cm
	\setlength\arrayrulewidth{1.0pt}
	\caption{Segmentation results on Mass-spring Hopper dataset.}
	\vspace{4pt}
	\begin{tabular}{lcccc}
		\toprule
     Method & NMI $\uparrow$ & ARI $\uparrow$ & Accuracy $\uparrow$ & $F_1$ $\uparrow$ \\
		\midrule
        Hybrid-SINDy  & 0.426 & 0.383 & 0.705 & 0.691 \cr
        AMORE (ours) &  \textbf{0.928} & \textbf{0.967} & \textbf{0.991} & \textbf{0.993} \cr
		\bottomrule
	\end{tabular}
	\label{Mass-spring Hopping Dataset segmentation table}
\end{table}

\begin{table}[t]
	\centering
        \footnotesize
	\tabcolsep=0.200cm
	\setlength\arrayrulewidth{1.0pt}
    \vspace{-1.5pt}
	\caption{Forecasting results of Location/Velocity on the Mass-spring Hopper dataset.}
	\vspace{4pt}
	\begin{tabular}{lcc}
		\toprule
     Method & NMAE $\downarrow$ & NRMSE $\downarrow$ \\
		\midrule
        LLMTime & 0.113 / 0.305 & 0.417 / 0.454 \cr
        TimeGPT & 0.092 / 0.217 &  0.322 / 0.340 \cr
        SVI  & 0.068 / 0.075 &  0.148 / 0.262  \cr
        Hybrid-SINDy & 0.240 / 0.314 & 0.336 / 0.372  \cr
        AMORE (ours) & \textbf{0.008} / \textbf{0.039} & \textbf{0.026} / \textbf{0.059}  \cr
		\bottomrule
	\end{tabular}
	\label{Mass-spring Hopping Dataset prediction table}
\end{table}

\begin{table}[t]
	\centering
        \footnotesize
	\tabcolsep=0.100cm
	\setlength\arrayrulewidth{1.0pt}
	\caption{Segmentation results on the SIR disease dataset.}
	\vspace{4pt}
	\begin{tabular}{lcccc}
		\toprule
     Method & NMI $\uparrow$ & ARI $\uparrow$ & Accuracy $\uparrow$ & $F_1$ $\uparrow$ \\
		\midrule
        Hybrid-SINDy  & 0.296 & 0.283 & 0.538 & 0.519 \cr
        AMORE (ours) &  \textbf{0.475} & \textbf{0.483} & \textbf{0.731} & \textbf{0.735} \cr
		\bottomrule
	\end{tabular}
	\label{SIR Disease Dataset segmentation table}
\end{table}
\begin{table}[t]
	\centering
        \footnotesize
	\tabcolsep=0.200cm
	\setlength\arrayrulewidth{1.0pt}
    \vspace{-2pt}
	\caption{Forecasting results of Susceptible/Infected on the SIR disease dataset.}
	\vspace{4pt}
	\begin{tabular}{lcc}
		\toprule
     Method & NMAE $\downarrow$ & NRMSE $\downarrow$ \\
		\midrule
        LLMTime & 0.352 / 0.396 & 0.481 / 0.523 \cr
        TimeGPT & 0.301 / 0.347 &  0.403 / 0.452 \cr
        SVI  & 0.257 / 0.273 &  0.355 / 0.401  \cr
        Hybrid-SINDy & 0.316 / 0.363 & 0.414 / 0.453  \cr
        AMORE (ours) & \textbf{0.088 / 0.113} & \textbf{0.142 / 0.181}  \cr
		\bottomrule
	\end{tabular}
	\label{SIR Disease Dataset prediction table}
\end{table}

\vspace{-5pt}
\paragraph{Implementation Details.}
We train all datasets with a fixed batch size of 40 for 20,000 training steps. We use the Adam optimizer with $10^{-5}$ weight-decay and clip gradients norm to 10. The learning rate is warmed up linearly from $5 \times 10^{-5}$ to $2 \times 10^{-4}$ for the first 2,000 steps, and then decays following a cosine manner with a rate of 0.99. Each experiment is running on one Nvidia GeForce RTX 3090 GPU. $d_{min}$ and $d_{max}$ of the count variables are simply set as 20 and 50, respectively for all datasets. The number of edge types $L$ is set as 2, containing one no-interaction type and one with-interaction type. More details are in Appendix~\ref{app: More Implementation Details}.

\vspace{-5pt}
\paragraph{Evaluation metrics.}
For evaluation of discovered equations, following~\citet{course2023state}, we use the reconstruction error between the discovered coefficients of equations and ground truth, i.e. ${\rm RER}=\frac{1}{T}\sum_{t=1}^T(||\mathbf{w}_t-\xi_t||_2\;/\;||\xi_t||_2)$ where $\mathbf{w}_t$ and $\xi_t$ are the learned and ground-truth coefficients at time $t$.
For evaluation of segmentation, following~\citet{ansari2021deep}, we use frame-wise segmentation accuracy, i.e. Accuracy and $F_1$ after matching the labels using the Hungarian algorithm~\citep{kuhn1955hungarian}, Normalized Mutual Information (NMI) and Adjusted Rand Index (ARI) to measure similarities. For evaluation of forecasting, we use Normalized Mean Absolute Error (NMAE) and Normalized Root Mean Squared Error (NRMSE).
We conducted each experiment with 5 random seeds. We report the average score of each experiment in the main paper and put the statistics (error bars) in the appendix due to limited space.

\vspace{-8pt}
\paragraph{Baselines.}
Hybrid-SINDy~\citep{mangan2019model} uses a two-stage paradigm and cannot perform forecasting, thus we compare with it on discovered equations and segmentation. 
To compare with Hybrid-SINDy on forecasting, we continue the value of the last observable time point as forecasting results of Hybrid-SINDy.
For forecasting, we compare with other three recent representative methods, i.e. SVI~\citep{course2023state} which is designed for equation discovery in non-hybrid systems and can perform forecasting, LLMTime~\citep{gruver2023large} which utilize pre-trained large language models (LLM) to do forecasting, and TimeGPT~\citep{garza2023timegpt} which is the first foundation model for time series.
GRASS~\citep{liu2023graph} does not discover equations, but models multi-object switching dynamical systems, so it is used for comparison in multi-object systems.

\subsection{Single-object Dynamical Systems}

\subsubsection{Mass-spring Hopper}
In the mass-spring hopper system, a mass and spring connect and hop on the ground with two modes, i.e. flight and compression. Details of the dataset are in Appendix~\ref{app: Mass-spring hopper}.
Comparison results of time series segmentation on the dataset are in Table~\ref{Mass-spring Hopping Dataset segmentation table}.
We can see that AMORE can achieve significant and consistent performance improvements across all metrics.
AMORE categorizes exactly two modes from the system and discovers equations for each mode
\begin{align*}
\begin{cases}
\dot{l} = v\;\;\textrm{and}\;\;\dot{v} = 11.03-10.08l \\
\dot{l} = v\;\;\textrm{and}\;\;\dot{v} = -1
\end{cases}
\end{align*}
which are nearly identical to the ground truth in Eq.~\eqref{eq: Mass-spring-scaling-final-for-generation}.
In Hybrid-SINDy, equations are discovered in each cluster fragment, thus producing a massive number of equations.
To quantitatively compare discovered equations, we compute $\rm RER$ for Hybrid-SINDy and AMORE which are $7.5e^{-3}$ and $2.4e^{-4}$, respectively.

Qualitative segmentation results of Hybrid-SINDy and AMORE are shown in Fig.~\ref{Qualitative time series segmentation results of AMORE compared to Hybrid-SINDy}.
Thanks to the amortized joint learning of modes and equations, AMORE can categorize the exact number of modes, achieve superior segmentation results, and discover high-quality equations. 
In these experiments, the maximal number of possible modes $K$ is set as 3 in our model. 
After learning, our model chooses 2 modes to be enough to categorize and express the dynamics of the specific hybrid systems.
Note that the number of discovered equations in Hybrid-SINDy is the same as the number of time points, which is much larger than the fixed number of modes, e.g. $K=3$ in our model.
To visualize discovered modes of Hybrid-SINDy, we aggregate the discovered equations with the same type of function terms as one mode, thus appearing more than 3 modes in the system.
Besides, ``AMORE w/o count'' represents our model without setting count variables. We can see that count variables can help AMORE learn fewer false-positive mode-switching behaviors. More quantitative ablation studies on count variables are in Appendix~\ref{app: Count Variables Analysis}.

We summarize time series forecasting results on the Mass-spring Hopper dataset in Table~\ref{Mass-spring Hopping Dataset prediction table}. We can see that our method significantly outperforms SVI which is designed for non-hybrid systems, and LLMTime as well as TimeGPT which utilizes pre-trained large models for forecasting, thanks to the proposed joint learning framework originally designed for hybrid systems.


\begin{table}[t]
	\centering
        \footnotesize
	\tabcolsep=0.150cm
	\setlength\arrayrulewidth{1.0pt}
	\caption{Forecasting results on non-hybrid dynamical systems. Results are shown in ${\rm log}_{10}({\rm NRMSE})$ where lower is better.}
	\vspace{4pt}
	\begin{tabular}{lccc}
		\toprule
     System & LLMTime & SVI & AMORE (ours) \\
		\midrule
        Coupled linear  & -0.39 &  -1.13 &  \textbf{-1.18}\cr
        Cubic oscillator & -0.45 &  -1.02  & \textbf{-1.06} \cr
        Lorenz'63 & -0.41 &  \textbf{-1.27} & -1.23 \cr
        Hopf bifurcation & -0.32 &  -0.94 & \textbf{-1.03}  \cr
        Selkov glycolysis & -0.68 & \textbf{-1.55}  & -1.49 \cr
        Duffing oscillator & -0.53 & -1.12  & \textbf{-1.17}  \cr
		\bottomrule
	\end{tabular}
	\label{Robustness to one-mode systems prediction table}
\end{table}

\subsubsection{SIR Disease Dataset}
The Susceptible, Infected and Recovered (SIR) disease model is an epidemiological model used to understand the spread of infectious diseases. Numbers of susceptible, infected, and recovered individuals are involved in model dynamics where some external events describe the modes, e.g. school in session or not. Detailed settings for this dataset are in Appendix~\ref{app: Susceptible, Infected and Recovered (SIR) Disease Dataset}.

We summarize the segmentation and forecasting results on the dataset in Tables~\ref{SIR Disease Dataset segmentation table} and~\ref{SIR Disease Dataset prediction table}. 
We can observe similar findings as in the Mass-spring Hopper dataset. AMORE can achieve consistently higher segmentation accuracy and lower forecasting errors across all metrics compared to Hybrid-SINDy, SVI, LLMTime, and TimeGPT.
AMORE categorizes exactly two modes from the system and discovers equations for each mode
\begin{align*}
\begin{cases}
\dot S \!=\! 2.74 \!-\! 0.0172\;IS \!-\! 0.0024\;S, \;\dot I \!=\! 0.0171\;IS \!-\! 0.2\;I\\
\dot S \!=\! 2.74 \!-\! 0.0057\;IS \!-\! 0.0021\;S, \;\dot I \!=\! 0.0051\;IS \!-\! 0.2\;I
\end{cases}
\end{align*}
which are nearly exact to the ground truth in Eq.~\eqref{eq: SIR-dataset-exact-number}.
Quantitative comparisons of the discovered equations are calculated by $\rm RER$ where Hybrid-SINDy and AMORE are $3.4e^{-3}$ and $1.8e^{-4}$, respectively.
We can see that AMORE can discover high-quality equations, and achieve superior segmentation and forecasting results thanks to the proposed joint learning framework designed for equation discovery in hybrid dynamical systems.

\subsubsection{Non-hybrid Dynamical Systems}
In some cases, we have prior knowledge of the dynamical systems whether they are hybrid or not. To answer the question of whether our method, which is originally designed for hybrid systems, can still perform well if we know the systems are non-hybrid in advance, we conduct experiments on six non-hybrid physical systems~\citep{course2023state}, including Coupled linear, Cubic oscillator, Lorenz'63, Hopf bifurcation, Selkov glycolysis, and Duffing oscillator.
Detailed settings of the datasets are in Appendix~\ref{app: Physical Systems}.
As we have the prior, we set the maximal possible number of modes in AMORE as 1 for all physical systems.
Following~\citet{course2023state}, we summarize the forecasting results in Table~\ref{Robustness to one-mode systems prediction table}. We can see that although our model is not specialized for non-hybrid systems, AMORE can still achieve better forecasting results on 4 out of 6 non-hybrid physical systems, which verifies the robustness of our model to non-hybrid dynamical systems.

\begin{figure*}[t!]
 \centering
 \setlength{\tabcolsep}{1pt}
 \includegraphics[width=.95\linewidth]{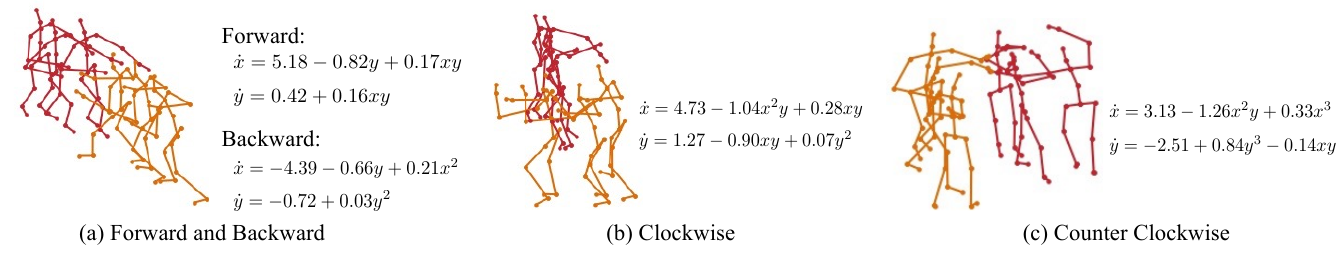}
 \vspace{-8pt}
 \caption{Discovered equations on the Salsa-dancing dataset. Locations $(x,y)$ of the hip joints are used as observations.}
 \label{Qualitative results of the discovered equations on the Salsa-dancing dataset}
 \end{figure*}

\subsection{Multi-object Hybrid Dynamical Systems}
Equation discovery in multi-object hybrid dynamical systems is an unexplored but more general setting.
In this section, we verify the effectiveness of the multi-object variant of our method, i.e. AMORE-MIO, on two multi-object datasets~\citep{liu2023graph}, i.e. the ODE-driven Particle dataset and the Salsa-dancing dataset.


\begin{table}[t]
	\centering
        \footnotesize
	\tabcolsep=0.100cm
	\setlength\arrayrulewidth{1.0pt}
    \vspace{-5pt}
	\caption{Segmentation results on ODE-driven Particle Dataset.}
	\vspace{4pt}
	\begin{tabular}{lcccc}
		\toprule
     Method & NMI $\uparrow$ & ARI $\uparrow$ & Accuracy $\uparrow$ & $F_1$ $\uparrow$ \\
		\midrule
        Hybrid-SINDy   & 0.205 & 0.192 & 0.414 & 0.407 \cr
        GRASS   & 0.445  & 0.437  & 0.732  & 0.726 \cr
        AMORE (ours)  & 0.418 & 0.405 & 0.692 &  0.684\cr
        AMORE-MIO (ours) & \textbf{0.453} & \textbf{0.442} & \textbf{0.741} & \textbf{0.735} \cr
		\bottomrule
	\end{tabular}
	\label{ODE-driven Particle Dataset segmentation table}
\end{table}

\begin{table}[t]
	\centering
        \footnotesize
	\tabcolsep=0.200cm
	\setlength\arrayrulewidth{1.0pt}
	\caption{Forecasting results of in terms of NMAE / NRMSE on ODE-driven Particle dataset.}
	\vspace{4pt}
	\begin{tabular}{lcc}
		\toprule
     Method & One-step & Multi-step  \\
        \midrule
        LLMTime & 0.335 / 0.438 & 0.370 / 0.473\cr
        TimeGPT & 0.351 / 0.445 & 0.392 / 0.490 \cr
        SVI  & 0.319 / 0.432 &  0.346 / 0.465   \cr
        Hybrid-SINDy & 0.340 / 0.431 &  0.372 / 0.487 \cr
        GRASS & 0.151 / 0.224  & 0.193 / 0.270 \cr
        AMORE (ours) & 0.184 / 0.265  & 0.217 / 0.302 \cr
        AMORE-MIO (ours) & \textbf{0.146 / 0.217} & \textbf{0.186 / 0.259} \cr
		\bottomrule
	\end{tabular}
	\label{ODE-driven Particle Dataset prediction table}
\end{table}
 
\subsubsection{ODE-driven Particle Dataset}
In ODE-driven particle systems, trajectories of particles are driven by Ordinary Differential Equations where particles switch their driven equations/modes when they collide with each other. 
Detailed settings of the ODE-driven Particle dataset are in Appendix~\ref{app: ODE-driven Particle Dataset}.
We summarize the segmentation results on the dataset in Table~\ref{ODE-driven Particle Dataset segmentation table}. 
We can see that our methods including both AMORE and AMORE-MIO achieve better time series segmentation results compared to Hybrid-SINDy and GRASS. Besides, AMORE-MIO can outperform AMORE consistently across all metrics.
AMORE-MIO categorizes 4 modes from the system and the discovered equations for each mode are
\begin{align*}
\begin{cases}
    \dot{x} = 1.08x - 0.92xy;\,\,\dot{y} = -0.93y + 1.11xy \notag\\
    \dot{x} = -0.17x^3 + 2.00y^3;\,\,\dot{y} = -2.13x^3 - 0.06y^3 \notag\\
    \dot{x} = 0;\,\,\dot{y} = 2.00 \notag\\
    \dot{x} = 0;\,\,\dot{y} = -2.00
\end{cases}
\end{align*}
which share the same number of coefficients and similar values as the ground truth in Eq.~\eqref{eq: ODE-driven Particle Dataset}. 
$\rm RER$ of discovered equations by Hybrid-SINDy, AMORE, and AMORE-MIO are $2.7e^{-2}$, $6.1e^{-3}$, and $4.3e^{-3}$, respectively, which shows that as a multi-object extension of AMORE, AMORE-MIO consistently outperforms AMORE and Hybrid-SINDy for equation discovery and mode categorization in multi-object hybrid systems thanks to the specially-designed interaction modeling of AMORE-MIO. We further show the forecasting results in Table~\ref{ODE-driven Particle Dataset prediction table}. We can see that AMORE-MIO consistently achieves the lowest forecasting errors for both one-step and multi-step predictions. Compared with GRASS, AMORE-MIO can obtain better results thanks to the introduced equation priors on the latent motion dynamics.

\begin{table}[t]
	\centering
        \footnotesize
	\tabcolsep=0.100cm
	\setlength\arrayrulewidth{1.0pt}
    \vspace{-5pt}
	\caption{Segmentation results on the Salsa-dancing dataset.}
	\vspace{4pt}
	\begin{tabular}{lcccc}
		\toprule
     Method & NMI $\uparrow$ & ARI $\uparrow$ & Accuracy $\uparrow$ & $F_1$ $\uparrow$ \\
		\midrule
        Hybrid-SINDy   & 0.102 & 0.097 & 0.325 & 0.309 \cr
        GRASS   & 0.173  & 0.177  & 0.579  & 0.526 \cr
        AMORE (ours)  & 0.167 & 0.173 & 0.565 & 0.518 \cr
        AMORE-MIO (ours) & \textbf{0.179} & \textbf{0.182} & \textbf{0.583}  & \textbf{0.531} \cr
		\bottomrule
	\end{tabular}
	\label{Salsa-dancing Dataset segmentation table}
\end{table}

\begin{table}[t]
	\centering
        \footnotesize
	\tabcolsep=0.200cm
	\setlength\arrayrulewidth{1.0pt}
    \vspace{-2pt}
	\caption{Forecasting results in terms of NMAE / NRMSE on the Salsa-dancing dataset.}
	\vspace{4pt}
	\begin{tabular}{lcc}
		\toprule
     Method & One-step & Multi-step \\
        \midrule
        LLMTime & 0.402 / 0.452 & 0.449 / 0.480 \cr
        TimeGPT & 0.341 / 0.417 &  0.394 / 0.446 \cr
        SVI  & 0.384 / 0.441 & 0.423 / 0.465  \cr
        Hybrid-SINDy & 0.362 / 0.405 & 0.416 / 0.433  \cr
        GRASS & 0.285 / 0.344 & 0.313 / 0.359 \cr
        AMORE (ours) & 0.291 / 0.361 & 0.334 / 0.373 \cr
        AMORE-MIO (ours) &\textbf{0.272 / 0.335}  & \textbf{0.301 / 0.352} \cr
		\bottomrule
	\end{tabular}
	\label{Salsa-dancing Dataset prediction table}
\end{table}

\begin{table}[t]
	\centering
        \footnotesize
	\tabcolsep=0.0600cm
	\setlength\arrayrulewidth{1.0pt}
	\caption{Analyses on robustness to different orders of polynomial as candidate basis functions on Mass-spring Hopper dataset. }
	\vspace{4pt}
	\begin{tabular}{lcccccc}
		\toprule
      Polynomial order & \multicolumn{2}{c}{2} & \multicolumn{2}{c}{3} & \multicolumn{2}{c}{5} \cr
     & NMI$\uparrow$ & RER$\downarrow$ & NMI$\uparrow$ & RER$\downarrow$ & NMI$\uparrow$ & RER$\downarrow$ \\
		\midrule
        Hybrid-SINDy   & 0.426 & $7.5e^{-3}$ & 0.384 & $8.1e^{-3}$ & 0.316 & $9.7e^{-3}$ \cr
            AMORE (ours)  & \textbf{0.934} & $\mathbf{2.1e^{-4}}$ & \textbf{0.936} & $\mathbf{2.3e^{-4}}$ & \textbf{0.933} & $\mathbf{2.8e^{-4}}$ \cr
		\bottomrule
	\end{tabular}
	\label{Analyses on robustness to different orders of polynomial}
\end{table}

\begin{table}[t]
	\centering
        \footnotesize
	\tabcolsep=0.0800cm
	\setlength\arrayrulewidth{1.0pt}
	\caption{Analyses on robustness to different maximal numbers of predefined modes on Mass-spring Hopper dataset.}
	\vspace{4pt}
	\begin{tabular}{lcccccc}
		\toprule
     Number of modes & \multicolumn{2}{c}{3} & \multicolumn{2}{c}{5} & \multicolumn{2}{c}{10} \cr
     & NMI$\uparrow$ & RER$\downarrow$ & NMI$\uparrow$ & RER$\downarrow$ & NMI$\uparrow$ & RER$\downarrow$ \\
		\midrule
            AMORE (ours)  & 0.934 & $2.1e^{-4}$ & 0.932 & $2.0e^{-4}$ & 0.937 & $2.1e^{-4}$ \cr
		\bottomrule
	\end{tabular}
	\label{Analyses on robustness to different maximal numbers of predefined modes}
\end{table}

\subsubsection{Salsa-dancing Dataset}
The Salsa-dancing dataset contains four modes, i.e. ``moving forward'', ``moving backward'', ``clockwise turning'', and ``counter-clockwise turning''.
Details of the dataset are in Appendix~\ref{app: Salsa-dancing Dataset}.
We summarize the segmentation and forecasting results on the Salsa-dancing dataset in Table~\ref{Salsa-dancing Dataset segmentation table} and Table~\ref{Salsa-dancing Dataset prediction table}. 
We observe similar findings in this real-world video dataset, as with the ODE-driven particle dataset. 
AMORE-MIO achieves significantly higher segmentation accuracies compared to Hybrid-SINDy and AMORE. 
Different from previous datasets, the salsa-dancing system is not generated synthetically by equations while results show that structural learning in the form of equations still benefits forecasting compared to purely autoregressive data-driven methods, i.e. LLMTime, SVI, and GRASS.
Qualitative results of the discovered equations on the dancing dataset are in Figure~\ref{Qualitative results of the discovered equations on the Salsa-dancing dataset}.

\subsection{Ablation Studies}
\paragraph{Sensitivity to order of polynomial functions.}
To test the sensitivity of our method to different orders of polynomials as candidate basis functions, we conduct experiments on the Mass-spring Hopper dataset by changing the order of polynomial functions to 2, 3, and 5. We present results in Table~\ref{Analyses on robustness to different orders of polynomial}. We observe that AMORE consistently outperforms Hybrid-SINDy, while AMORE is not sensitive to the polynomial orders compared to Hybrid-SINDy.

\vspace{-5pt}
\paragraph{Sensitivity to number of dynamic modes}
We test the robustness of our method to different maximum numbers of modes, that is 3, 5, and 10, while the true number is 2 on the Mass-spring Hopper dataset. The results of segmentation and discovered equations are in Table~\ref{Analyses on robustness to different maximal numbers of predefined modes}. We can see that AMORE is impervious to this misspecification, which indicates that we can set a large number of possible modes while AMORE can still learn those needed.

\vspace{-5pt}
\paragraph{Sensitivity to more complex dynamical systems}
We originally followed the setup of Hybrid-SINDy, where all of the dynamics can be approximated by polynomial basis functions. However, our model is not limited to these functions. To show results on more complex dynamical systems, we conduct experiments on a synthetic dataset where two modes are driven by $\dot{x} = x+x^2+cos(x)$ and $\dot{x} = x+e^x$, respectively. We set the basis functions as polynomials order 3 together with $\{cos(x), sin(x),  e^x\}$. The discovered equations by our model are
\begin{align*}
\begin{cases}
\dot{x} = 0.97x+1.02x^2+1.08cos(x) \\
\dot{x} =0.05+1.12x+0.96e^x
\end{cases}
\end{align*}
When we set the basis functions as polynomials order 3 without $\{cos(x), sin(x),  e^x\}$. The discovered equations by our model are 
$\dot{x} = 0.92+x+0.76x^2$ and $\dot{x} = 1.26+1.31x+0.83x^2+0.34x^3$, respectively.
We can see that our model can be extended to equation discovery with more complex basis functions. When the candidate basis functions are limited to polynomial functions, our model can discover approximated ones with more terms, complexity, and errors.

\paragraph{Model complexity analysis}
The numbers of parameters used in baseline methods are summarized in Table~\ref{Comparisons on model complexity regarding the numbers of learnable parameters table}. As Hybrid-SINDy is not a deep learning method and does not use neural networks, it does not involve learnable parameters and does not need much data for the training of any little parameters the model has. This comes at the expense that Hybrid-SINDy tends to not generalize beyond simple dynamical settings, as shown in our experiments in the main paper. When given a complex dynamical setting with sufficient data, AMORE and AMORE-MIO perform better and have slightly fewer parameters than the other deep learning-based approaches, except LLMTime.

\begin{table}[t]
	\centering
        \footnotesize
	\tabcolsep=0.100cm
	\setlength\arrayrulewidth{1.0pt}
	\caption{Comparisons on model complexity regarding the numbers of learnable parameters.}
	\vspace{4pt}
	\begin{tabular}{lc}
		\toprule
     Method & Number of parameters \\
		\midrule
       Hybrid-SINDy  & 0 \cr
       AMORE (ours) &  2,240 \cr
       AMORE-MIO (ours) &  2,512 \cr
       GRASS &  4,628 \cr
       SVI & 2,826 \cr
       LLMTime & 175 billion (GPT-3) \cr
		\bottomrule
	\end{tabular}
	\label{Comparisons on model complexity regarding the numbers of learnable parameters table}
\end{table}

\section{Conclusion and Future work}
In this paper, we reformulate the problem of equation discovery in hybrid dynamical systems and propose an end-to-end learning framework, i.e. Amortized Equation Discovery (AMORE) to jointly categorize motion dynamics and discover equations by modeling categorical modes and mode-switching behaviors. 
Besides, we extend our method to multi-object scenarios, i.e. AMORE-MIO, which is unexplored by previous methods and a more natural setting. Extensive experiments on 10 hybrid and non-hybrid systems demonstrate the effectiveness of our method. Future work can include equation discovery with partial known knowledge, equation discovery from videos of hybrid systems, and more complex candidate basis functions.

\newpage
\section*{Impact Statement}
This paper presents work whose goal is to advance the field of Machine Learning and Dynamical Systems. There are many potential societal consequences of our work, none of which we feel must be specifically highlighted here.

\section*{Acknowledgements}
This work is financially supported by NWO TIMING VI.Vidi.193.129.
We also thank SURF for the support in using the National Supercomputer Snellius.

\bibliography{icml_paper}
\bibliographystyle{icml2024}

\newpage
\appendix
\onecolumn
{\Large \textbf{Appendix}}
\section{More Details of AMORE}
\subsection{Neural Network Implementation}
\label{app: Neural Network Implementation of AMORE}
We use neural networks to model the joint generative probabilities of hybrid systems in our model, i.e. Eq.~\eqref{eq:joint-full-single-object}. For the initial states, we model the initial prior distributions as:
\begin{align*}
p(z_1) &= \rm{Cat}(z_1; \bm{\pi}), \\
p(\mathbf{y}_1|z_1) &= \mathcal{N}(\mathbf{y}_1;\bm{\mu}_{z_1};\bm{\Sigma}_{z_1}),
\end{align*}
where $\rm{Cat}$ and $\mathcal{N}$ denote categorical and multivariate Gaussian distributions, respectively. We set the prior distribution of $p(z_1)$ as uniform to encourage diversity.

\paragraph{Count variables and count transition probability.} To implement the count variables, we set a categorical distribution over $\{d_{\rm min}, \cdots, d_{\rm max}\}$ for each mode, where $d_{\rm min}$ and $d_{\rm max}$ are the minimal and maximal numbers of time steps before making a mode switch. The count transition probability $p(c_t|c_{t-1},z_{t-1})$ is modeled as a learnable matrix $\mathbf{P} \in \mathbb{R}^{K\times (d_{\rm max}-d_{\rm min}+1)}$, which is fixed across all time steps. Each term $\rho_k(c)$ in $\mathbf{P}$ represents the probability of the $k$-th mode switching to another mode when its current count is $c$. The probability of a count increment at count $c$ for mode $k$ can be calculated as
\begin{align*}
\mu_k(c) = 1 - \frac{\rho_k(c)}{\sum_{d=c}^{d_{\rm max}}\rho_k(d)}.
\end{align*}
The count transition probability is thus defined as
\begin{align*}
\!\! \!  p(c_{t}|c_{t-1}, z_{t-1}\!=\!k)\!=\!\!
\begin{cases}\!
\mu_k(c_{t-1})    &  \!\!\!\! {\rm if}\;c_{t}\!=\!c_{t-1}\!+ 1\!\\
1\!-\!\mu_k(c_{t-1})  & \!\!\!\!{\rm if}\;c_{t} = 1
\! \end{cases}.
\end{align*}

\paragraph{Mode variables and mode transition probability.} Since the mode variables $z_t$ take one out of $K$ possible values, we model them as categorical variables, parameterized by mode transition matrix $\mathbf{T}_t$ at timestep $t$. The mode transition probability is modeled as
\begin{align*}
\!\! \!  p(z_{t}|z_{t-1}, c_{t}, \mathbf{y}_{t-1}\!)\!=\!
\begin{cases}
\delta_{z_{t}=z_{t-1}}    &  \!\!\!\! {\rm if}\;c_{t} > 1 \\
{\rm Cat}(z_{t};  \mathbf{T}_t )  & \!\!\!\!{\rm if}\;c_{t} = 1
\end{cases},
\end{align*}
where we resample the modes or preserve them depending on whether count variables are reset to 1 or not. We model the parameters $\mathbf{T}_t$ of the categorical distributions with a neural network, i.e. a simple MLP, $\mathbf{T}_t = f_z(\mathbf{y}_{t-1})$ that takes as input the observations. The network returns a $K \times K$ transition matrix per time step $t$, where rows correspond to past modes $z_{t-1}$ and columns current modes $z_t$. Each term $\tau_t^{j,k}$ in $\mathbf{T}_t$ represents the probability of mode $j$ switching to mode $k$ at timestep $t$. To satisfy the positivity $\tau_t^{j,k} > 0, \; \forall j,k=1,\cdots,K$ and $\ell_1$ constraints $\sum_{k} \tau_t^{j,k}=1, \; \forall j=1,..., K$, we apply a tempered softmax after $f_z$, i.e. $\mathcal{S}_{\tau_z} \circ f_z(\cdot)$.

\subsection{Inference Model of AMORE}
\label{app: Inference Model of AMORE}
We perform conditionally exact inference for the two discrete latent variables, i.e. modes $\mathbf{z}_{1:T}$ and counts $\mathbf{c}_{1:T}$, similar to the forward-backward procedure for HMM~\citep{eddy1996hidden}.
Conditioned on observations $\mathbf{y}_{1:T}$, the posterior joint distribution $p_\theta(\mathbf{z}_{1:T},\mathbf{c}_{1:T}|\mathbf{y}_{1:T})$ is calculated by modifying the forward-backward recursions to handle the joint hierarchical latent variables. Specifically, the forward $\alpha_t$ and backward $\beta_t$ parts are defined as
\begin{align*}
\alpha_t(z_t,c_t) &= p(z_t,c_t,\mathbf{y}_{1:t}), \\
\beta_t(z_t,c_t) &= p(\mathbf{y}_{t+1:T}|\mathbf{y}_{t},z_t,c_t).
\end{align*}
Specifically, the posterior joint probability of mode and count variables $\mathbf{z}$, $\mathbf{c}$ conditioned on observations $\mathbf{y}$ is calculated as
\begin{align*}
    p(z_t,c_t|\mathbf{y}_{1:T}) & \propto p(z_t, c_t,\mathbf{y}_{1:T}) \\
    &= \underbrace{p(z_t,c_t,\mathbf{y}_{1:t})}_{Forward}\underbrace{p(\mathbf{y}_{t+1:T}|\mathbf{y}_{t},z_t,c_t)}_{Backward} \\
    &= \alpha_t(z_t,c_t)\cdot\beta_t(z_t,c_t).
\end{align*}
The derivatives of the forward section $\alpha_t(z_t,c_t)$ are
\begin{align*}
    \alpha_1(z_1,c_1) 
    &= p(z_1, c_1, \mathbf{y}_1) \\
    &= \delta_{c_1=1} p(z_1)p(\mathbf{y}_1|z_1), \\
    \underline{\alpha_t(z_t, c_t)} &= p(z_t,c_t,\mathbf{y}_{1:t}) \\
    &= \sum_{z_{t-1},c_{t-1}}p(z_t,c_t,\mathbf{y}_{1:t},z_{t-1},c_{t-1}) \\
    &= \sum_{z_{t-1},c_{t-1}}p(z_{t-1},c_{t-1},\mathbf{y}_{1:t-1})p(c_{t}|c_{t-1},z_{t-1})p(z_{t}|z_{t-1},c_{t},\mathbf{y}_{t-1})p(\mathbf{y}_{t}|\mathbf{y}_{t-1},z_{t}) \\
    &= p(\mathbf{y}_{t}|\mathbf{y}_{t-1},z_{t})\sum_{z_{t-1},c_{t-1}}\underline{\alpha_{t-1}(z_{t-1},c_{t-1})}p(c_{t}|c_{t-1},z_{t-1})p(z_{t}|z_{t-1},c_t,\mathbf{y}_{t-1})\\
    &= p(\mathbf{y}_{t}|\mathbf{y}_{t-1},z_{t})\Bigg[\delta_{c_t=1}\sum_{z_{t-1}}p(z_t|z_{t-1},c_t,\mathbf{y}_{t-1})\sum_{c_{t-1}}(1-\mu_{z_{t-1}(c_{t-1})})\alpha_{t-1}(z_{t-1},c_{t-1}) \\
    & + \delta_{\substack{z_{t-1} = z_{t} \\ c_t>1 \\ c_{t-1} = c_{t} - 1}}\mu_{z_{t-1}}(c_{t-1})\alpha_{t-1}(z_{t-1},c_{t-1}) \Bigg],
\end{align*}
where $\alpha_t(z_t,c_t)$ can be expressed by $\alpha_{t-1}(z_{t-1},c_{t-1})$ recursively with states transitions.

The derivatives of the backward section $\beta_t(z_t,c_{t})$ are
\begin{align*}
    \beta_T(z_T,c_T) &= 1 \\
    \underline{\beta_t(z_t,c_t)} &= p(\mathbf{y}_{t+1:T}|\mathbf{y}_{t},z_t,c_t) \\
    &= \sum_{z_{t+1},c_{t+1}}p(\mathbf{y}_{t+1:T},z_{t+1},c_{t+1} |\mathbf{y}_{t}, z_{t}, c_{t})\\
    &= \sum_{z_{t+1},c_{t+1}} p(c_{t+1}|c_{t},z_{t})p(z_{t+1}|z_{t},c_{t},\mathbf{y}_{t})p(\mathbf{y}_{t+1}|\mathbf{y}_{t},z_{t+1})p(\mathbf{y}_{t+2:T}|\mathbf{y}_{t+1}, z_{t+1}, c_{t+1}) \\
    &= \sum_{z_{t+1},c_{t+1}} p(c_{t+1}|c_{t},z_{t})p(z_{t+1}|z_{t},c_{t+1},\mathbf{y}_{t})p(\mathbf{y}_{t+1}|\mathbf{y}_{t},z_{t+1})\underline{\beta_{t+1}(z_{t+1},c_{t+1})} \\
    &= \delta_{\substack{c_{t+1}=1 \\ c_t \geq d_{\rm min}}}(1-\mu_{z_t}(c_t))\sum_{z_{t+1}}p(z_{t+1}|z_{t},c_{t+1},\mathbf{y}_{t})p(\mathbf{y}_{t+1}|\mathbf{y}_{t},z_{t+1})\beta_{t+1}(z_{t+1},c_{t+1}) \\
    &+ \delta_{\substack{c_{t+1} = c_t +1 \\ z_{t+1}=z_t}}\mu_{z_t}(c_t)p(\mathbf{y}_{t+1}|\mathbf{y}_{t},z_{t+1})\beta_{t+1}(z_{t+1},c_{t+1}),
\end{align*}
where $\beta_t(z_t,c_t)$ can be computed via $\beta_{t+1}(z_{t+1},c_{t+1})$ recursively with states transitions.

\subsection{Derivation of Optimization Objective}
\label{app: Derivation of Training Objective of AMORE}
The optimization objective of our model is to maximize the observation likelihood ${\rm log}\,p(\mathbf{y})$ with sparse regularization on coefficients of candidate basis functions, where the observation likelihood ${\rm log}\,p(\mathbf{y})$ can be calculated as
\begin{align*}
\rm log\,p(\mathbf{y}) 
&= \mathbb{E}_{p(\mathbf{z},\mathbf{c}|\mathbf{y})}\left[ {\rm log}\,p(\mathbf{y})\right] \\
&= \mathbb{E}_{p(\mathbf{z},\mathbf{c}|\mathbf{y})}\left[ {\rm log}\,p(\mathbf{y},\mathbf{z},\mathbf{c})\right] - \mathbb{E}_{p(\mathbf{z},\mathbf{c}|\mathbf{y})}\left[ {\rm log}\,p(\mathbf{z},\mathbf{c}|\mathbf{y})\right] \\
&= \mathbb{E}_{p(\mathbf{z},\mathbf{c}|\mathbf{y})}\left[ {\rm log}\,p(\mathbf{y},\mathbf{z},\mathbf{c})\right],
\end{align*}
where $\mathbb{E}_{p(\mathbf{z},\mathbf{c}|\mathbf{y})}\left[ {\rm log}\,p(\mathbf{z},\mathbf{c}|\mathbf{y})\right]$ is calculated as 
\begin{align*}
\mathbb{E}_{p(\mathbf{z},\mathbf{c}|\mathbf{y})}\left[ {\rm log}\,p(\mathbf{z},\mathbf{c}|\mathbf{y})\right] = \int p(\mathbf{z},\mathbf{c}|\mathbf{y})\frac{ {\rm log}\,p(\mathbf{z},\mathbf{c}|\mathbf{y})}{p(\mathbf{z},\mathbf{c}|\mathbf{y})}d(\mathbf{z},\mathbf{c}) =  \int{\rm log}\,p(\mathbf{z},\mathbf{c}|\mathbf{y})d(\mathbf{z},\mathbf{c}) =   1 = 0.
\end{align*}
Following Markovian property, we expand $ {\rm log}\,p(\mathbf{y},\mathbf{z},\mathbf{c})$ over time and calculate it as
\begin{align*}
     {\rm log}\,p(\mathbf{y},\mathbf{z},\mathbf{c}) &=  \,{\rm log}\,p(\mathbf{y}_{1:T},\mathbf{z}_{1:T},\mathbf{c}_{1:T}) \\
    &=  \,{\rm log}\!\left[p(\mathbf{y}_{1}|z_{1})p(z_{1})\right] + \sum_{t=2}^{T} \,{\rm log}\!\left[p(\mathbf{y}_{t}|\mathbf{y}_{t-1},z_{t})p(z_{t}|z_{t-1},c_{t},\mathbf{y}_{t-1})p(c_{t}|c_{t-1},z_{t-1})\right].
\end{align*}
Finally, combined with expectations, ${\rm log}\,p(\mathbf{y})$ can be calculated as
\begin{align*}
      {\rm log}\,p(\mathbf{y})&=\mathbb{E}_{p(\mathbf{z},\mathbf{c}|\mathbf{y})}\left[  {\rm log}\,p(\mathbf{y},\mathbf{z},\mathbf{c})\right],\\
    &= \mathbb{E}_{p(\mathbf{z}_{1:T},\mathbf{c}_{1:T}|\mathbf{y}_{1:T})}\left[  {\rm log}\,p(\mathbf{y}_{1:T},\mathbf{z}_{1:T},\mathbf{c}_{1:T})\right] \\
    &= \sum_{k}p(z_1=k|\mathbf{y}_{1:T})   \,{\rm log}\left[p(\mathbf{y}_{1}|z_{1})p(z_{1}=k)\right] \\
    &\,\,\,\,\,\,\,+ \sum_{t=2}^{T}\sum_{k,j,u,v}\!\! \xi(k,j,u,v)\,  \,{\rm log}\!\left[p(\mathbf{y}_{t}|\mathbf{y}_{t-1},z_{t}=k)p(z_{t}\!=\!k|z_{t-1}\!=\!j, c_{t}\!=\!v,\mathbf{y}_{t-1})p(c_{t}\!=\!v|c_{t-1}\!=\!u,z_{t-1}\!=\!j)\right]\\
    &= \sum_{k}\gamma(k)\,   \,{\rm log}\!\left[B_1(k)\cdot \pi(k)\right] \\
    &\,\,\,\,\,\,\,+\sum_{t=2}^{T}\sum_{k,j,u,v}\!\!\xi(k,j,u,v)\,  \,{\rm log}\!\left[B_t(k)\cdot 
    A_t(k,j,v) \cdot C_t(j,u,v)\right]
\end{align*}
where $\pi(k)$, $\gamma(k)$, $\xi(k,j,u,v)$, $B_t(k)$, $A_t(k,j,v)$, and $C_t(j,u,v)$ are defined as
\begin{align*}
\pi(k) &= p(z_{1}=k),\\
\gamma(k) &= p(z_1=k|\mathbf{y}_{1:T}),\\
\xi(k,j,u,v) &= p(z_t\!\!=\!k,z_{t-1}\!\!=\!j,c_t\!=\!v,c_{t-1}\!=\!u|\mathbf{y}_{1:T}), \\
B_t(k) &= p(\mathbf{y}_{t}|\mathbf{y}_{t-1},z_{t}=k),\\
A_t(k,j,v) &= p(z_{t}\!=\!k|z_{t-1}\!=\!j, c_{t}\!=\!v,\mathbf{y}_{t-1}),\\
C_t(j,u,v) &= p(c_{t}\!=\!v|c_{t-1}\!=\!u,z_{t-1}\!=\!j).
\end{align*}
$\pi(\mathbf{k})$ is the initial discrete mode probability. $B_t(k)$ is the continuous state transition probability conditioned on different types of discrete modes $k$. $A_t(k,j,v)$ is the discrete mode transition probability. $C_t(j,u,v)$ is the mode duration count transition probability. Besides, $\gamma(k) = p(z_1=k|\mathbf{y}_{1:T})$ and $\xi(k,j,u,v) = p(z_t\!\!=\!k,z_{t-1}\!\!=\!j,c_t\!=\!v,c_{t-1}\!=\!u|\mathbf{y}_{1:T})$ can be calculated similarly to the forward and backward algorithm in HMMs~\citep{eddy1996hidden} which is detailed in Appendix~\ref{app: Inference Model of AMORE}.

\section{More Details of AMORE-MIO}
\subsection{Expansion of Generative Model over Objects}
\label{app: Expansion of Generative Model over Objects of AMORE-MIO}
The joint generative probability of AMORE-MIO for multi-object hybrid systems is expanded over objects as
\begin{align*}
p(\mathbf{y}, \mathbf{z}, \mathbf{c}, \mathbf{e}) &=
\underbrace{\prod_{n=1}^N \!p(\mathbf{y}_1^n|z_1^{n}) \cdot \prod_{n=1}^N \!p(z_1^n) \cdot \prod_{n=1}^{N}\prod_{m=1}^{N}\!p(e_1^{m\rightarrow n})}_{\rm{Initial\,\,states}} \cdot \prod_{t=2}^{T} \Bigg[ \prod_{n=1}^{N} p(\mathbf{y}_{t}^{n}|\mathbf{y}_{t-1}^{n},z_{t}^{n}) \cdot \prod_{n=1}^{N} p(c_{t}^{n}|c_{t-1}^{n},z_{t-1}^{n}) \cdot \\
&\;\;\;\;\;\;\prod_{n=1}^{N}\!\sum_{m=1}^{N}p(z_{t}^{n}|z_{t-1}^{m}, c_{t}^{n}, e_{t}^{m\rightarrow n}, \mathbf{y}_{t-1}^{m}, \mathbf{y}_{t-1}^{n}) \cdot
\prod_{n=1}^{N}\!\sum_{m=1}^{N}p(e_{t}^{m\rightarrow n}|e_{t-1}^{m\rightarrow n}, \mathbf{v}_{t-1}^m, \mathbf{v}_{t-1}^n)\Bigg],
\end{align*}
where in the initial states, we model for each object $n$ an initial mode and observation distributions, i.e. $p(z_1^n)$ and $p(\mathbf{y}_1^n|z_1^{n})$. For each pair of interactions, $p(e_1^{m\rightarrow n})$ models the initial edge distribution. For later time steps $t \geq 2$, $p(e_{t}^{m\rightarrow n}|e_{t-1}^{m\rightarrow n}, \mathbf{v}_{t-1}^m, \mathbf{v}_{t-1}^n)$ models the edge variable transition probability conditioned on node states $\{\mathbf{v}_{t-1}^m, \mathbf{v}_{t-1}^n\}$ in graph $\mathcal{G}_t$. $p(z_{t}^{n}|z_{t-1}^{m}, c_{t}^{n}, e_{t}^{m\rightarrow n}, \mathbf{y}_{t-1}^{m,n})$ models how the modes of objects are affected by the modes of all other objects, conditioned on count variables $c_{t}^{n}$, edge variables $e_{t}^{m\rightarrow n}$, and observations $\{\mathbf{y}_{t-1}^{m}, \mathbf{y}_{t-1}^{n}\}$. $p(\mathbf{y}_{t}^{n}|\mathbf{y}_{t-1}^{n},z_{t}^{n})$ and $p(c_{t}^{n}|c_{t-1}^{n},z_{t-1}^{n})$ model for each object an observation transition probability and count variable transition probability.

\subsection{Neural Network Implementation}
\label{app: Neural Network Implementation of AMORE-MIO}
Implementations of $p(\mathbf{y}_1^n|z_1^{n})$, $p(z_1^n)$, $p(\mathbf{y}_{t}^{n}|\mathbf{y}_{t-1}^{n},z_{t}^{n})$, and $p(c_{t}^{n}|c_{t-1}^{n},z_{t-1}^{n})$ in multi-object scenarios are the same as those in single-object scenarios. Next, we elaborate on how we implement the other terms, i.e. $p(e_1^{m\rightarrow n})$, $p(e_{t}^{m\rightarrow n}|e_{t-1}^{m\rightarrow n}, \mathbf{v}_{t-1}^m, \mathbf{v}_{t-1}^n)$, and $p(z_{t}^{n}|z_{t-1}^{m}, c_{t}^{n}, e_{t}^{m\rightarrow n}, \mathbf{y}_{t-1}^{m}, \mathbf{y}_{t-1}^{n})$.

\paragraph{Edge variables and edge transition probability.} We implement the edge variable $\mathbf{e}$ as a categorical distribution over $\{1, \cdots, L\}$ for $L$ possible interaction types including a \emph{no-interaction} type. We set the prior distribution to be higher for \emph{no-interaction} edges in $p(e_1^{m\rightarrow n})$ to encourage sparse graphs. The edge transition probability is modeled as
\begin{align*}
p(e_{t}^{m\rightarrow n}|e_{t-1}^{m\rightarrow n}, \mathbf{v}_{t-1}^m, \mathbf{v}_{t-1}^n) = {\rm Cat}(e_t^{m\rightarrow n}; \mathcal{S}_{\tau_e}(f_e(e_{t-1}^{m\rightarrow n}, \mathbf{v}_{t-1}^m, \mathbf{v}_{t-1}^n))),
\end{align*}
where The neural network $f_e$ takes $e_{t-1}^{m\rightarrow n}$, $\mathbf{v}_{t-1}^m$, and $\mathbf{v}_{t-1}^n$ as input and outputs the probabilities of all possible edge types at time step $t$, which are further post-processed by a tempered softmax function $\mathcal{S}_{\tau_e}$ with temperature $\tau_e$ to ensure normalization. In practice, the edge transition network $f_e$ is a single hidden layer MLP.

\paragraph{Extension of mode transition probability.} After getting $e_{t}^{m\rightarrow n}$ by the edge transition probability, we show how $e_{t}^{m\rightarrow n}$ affects the mod-switching behaviors. We model the mode transition probability in multi-object hybrid systems as
\begin{align*}
& p(z_{t}^n|z_{t-1}^m, c_{t}^n, e_t^{m \rightarrow n}, \mathbf{y}_{t-1}^m, \mathbf{y}_{t-1}^n\!) =\! \begin{cases}
\delta_{z_{t}^n=z_{t-1}^n}    &  \!\!\!\! {\rm if}\;c_{t}^n \!>\! 1 \\
{\rm Cat}(z_{t}^n; \mathcal{S}_{\tau_z}(\sum_{l} e_{t,l}^{m\rightarrow n} f_l(\mathbf{y}_{t-1}^m, \mathbf{y}_{t-1}^n))  & \!\!\!\!{\rm if}\;c_{t}^n \!=\! 1
\end{cases}\!,
\end{align*}
where $\delta$ and $\mathcal{S}_{\tau_z}$ are a Kronecker function and a tempered softmax function. $e_{t,l}^{m\rightarrow n}$ denotes the probability of each edge type $l$. We set a neural network $f_l$ for each edge type $l$ (totally $L$) to model different interaction effects, which are normalized by $e_{t,l}^{m\rightarrow n}$ to aggregate effects from all the interaction types.

\subsection{Inference Model of AMORE-MIO}
\label{app: Inference Model of AMORE-MIO}
\paragraph{Approximate inference of edge variables.}We use a graph neural network $f_{\phi_e}(\mathbf{y})$ to conduct approximate inference of edge variables $\mathbf{e}$, i.e. $q_{\phi_e}(\mathbf{e}|\mathbf{y})$. The node embeddings in the latent graph $\mathcal{G}_t$ are the observations $\mathbf{y}$, and the edge embeddings are calculated by two rounds of message-passing
\begin{align*}
                      &\mathbf{h}_{n}^1 = f_{\phi_e}^{\rm emb}(\mathbf{y}_t^{n}), \\
v\rightarrow e\!: \,\,&\mathbf{h}_{m\rightarrow n}^1 = f_{\phi_z}^{e,1}([\mathbf{h}_{m}^1, \mathbf{h}_{n}^1]), \\
e\rightarrow v\!: \,\,&\mathbf{h}_{n}^2 = f_{\phi_e}^{v,1}(\sum_{m=1}^N \mathbf{h}_{m\rightarrow n}^1), \\
v\rightarrow e\!: \,\,&\mathbf{h}_{m\rightarrow n}^2 = f_{\phi_e}^{e,2}([\mathbf{h}_{m}^2, \mathbf{h}_{n}^2]),
\end{align*}
where $\mathbf{h}_{m\rightarrow n}^2$ is further processed by a tempered Gumbel softmax ${\rm softmax}((\mathbf{h}_{m\rightarrow n}^2 + \mathbf{g})/\tau)$ to achieve $q_{\phi_e}(\mathbf{e}|\mathbf{y})$, to be more specific $q_{\phi_e}(e_t^{m\rightarrow n}|\mathbf{y}_t^m,\mathbf{y}_t^n)$. Here, we use continuous relaxation and reparameterization of discrete distributions for gradient backpropagation~\cite{kipf2018neural}. $\mathbf{g}$ is a vector sampled from a ${\rm Gumbel}(0,1)$ distribution and the softmax temperature $\tau$ controls relaxation smoothness.

\paragraph{Exact inference of mode and count variables.} Given the approximate edge variables $\tilde{\mathbf{e}}\sim q_{\phi_e}\!(\mathbf{e}|\mathbf{y})$, we do exact inference of the mode and count variables $p_\theta(\mathbf{z},\mathbf{c}|\mathbf{y},\tilde{\mathbf{e}})$. Similar to the single-object scenarios, the conditional joint distribution is calculated by modifying the forward-backward algorithm. Specifically, the forward $\alpha_t$ and backward $\beta_t$ are calculated as
\begin{align*}
\!\alpha_t(\mathbf{z}_t^{1:N},\mathbf{c}_t^{1:N})\! &= \!p(\mathbf{z}_t^{1:N},\mathbf{c}_t^{1:N},\mathbf{y}_{1:t}^{1:N},\mathbf{e}_{1:t}^{1:N^2}), \\
\!\beta_t(\mathbf{z}_t^{1:N},\mathbf{c}_t^{1:N})\! &= \!p(\mathbf{y}_{t+1:T}^{1:N}|\mathbf{y}_{t}^{1:N},\mathbf{z}_t^{1:N},\mathbf{c}_t^{1:N},\mathbf{e}_{t}^{1:N^2}).
\end{align*}

Specifically, the joint probability of mode and count variables $\mathbf{z}$, $\mathbf{c}$ conditioned on observations $\mathbf{y}$ and approximate edge variables $\mathbf{e}$ is calculated as
\begin{align*}
    p(\mathbf{z}_t,\mathbf{c}_t|\mathbf{y}_{1:T},\mathbf{e}_{1:T}) & \propto p(\mathbf{z}_t,\mathbf{c}_t,\mathbf{y}_{1:T},\mathbf{e}_{1:T}) \\
    &= \underbrace{p(\mathbf{z}_t,\mathbf{c}_t,\mathbf{y}_{1:t},\mathbf{e}_{1:t})}_{Forward}\underbrace{p(\mathbf{y}_{t+1:T},\mathbf{e}_{t+1:T}|\mathbf{y}_t,\mathbf{z}_t,\mathbf{c}_t}_{Backward}) \\
    &= \alpha_t(\mathbf{z}_t,\mathbf{c}_t)\cdot\beta_t(\mathbf{z}_t,\mathbf{c}_t).
\end{align*}
The derivatives of the forward section $\alpha_t(\mathbf{z}_t,\mathbf{c}_t)$ is calculated as:
\begin{align*}
    \alpha_1(\mathbf{z}_1, \mathbf{c}_1) 
    &= p(\mathbf{z}_1, \mathbf{c}_1, \mathbf{y}_1, \mathbf{e}_1) \\
    &= 
    p(\mathbf{z}_1^{1:N}, \mathbf{c}_1^{1:N}, \mathbf{y}_1^{1:N}, \mathbf{e}_1^{1:N^2})\\
    &= \delta_{\mathbf{c}_1^{1:N}=1} p(\mathbf{z}_1^{1:N})p(\mathbf{e}_1^{1:N^2})p(\mathbf{y}_{1}^{1:N}|\mathbf{z}_1^{1:N}) \\
    &=  \delta_{\mathbf{c}_1^{1:N}=1} p(\mathbf{z}_1^{1:N}) p(\mathbf{e}_1^{1:N^2})\prod_{n=1}^{N}p(\mathbf{y}_{1}^{n}|\mathbf{z}_{1}^{n})\\
    \underline{\alpha_t(\mathbf{z}_t, \mathbf{c}_t)} &= p(\mathbf{z}_t,\mathbf{c}_t,\mathbf{y}_{1:t},\mathbf{e}_{1:t}) \\
    &= p(\mathbf{z}_t^{1:N},\mathbf{c}_t^{1:N},\mathbf{y}_{1:t}^{1:N},\mathbf{e}_{1:t}^{1:N^2})\\
    &= \sum_{\mathbf{z}_{t-1}^{1:N},\mathbf{c}_{t-1}^{1:N}}p(\mathbf{z}_t^{1:N},\mathbf{c}_t^{1:N},\mathbf{y}_{1:t}^{1:N},\mathbf{e}_{1:t}^{1:N^2}, \mathbf{z}_{t-1}^{1:N},\mathbf{c}_{t-1}^{1:N}) \\
    &= \sum_{\mathbf{z}_{t-1}^{1:N},\mathbf{c}_{t-1}^{1:N}}\Bigg[p(\mathbf{z}_{t-1}^{1:N},\mathbf{c}_{t-1}^{1:N},\mathbf{y}_{1:t-1}^{1:N},\mathbf{e}_{1:t-1}^{1:N^2})p(\mathbf{y}_{t}^{1:N}|\mathbf{y}_{t-1}^{1:N},\mathbf{z}_{t}^{1:N})p(\mathbf{z}_{t}^{1:N}|\mathbf{z}_{t-1}^{1:N},\mathbf{c}_{t}^{1:N},\mathbf{y}_{t-1}^{1:N},\mathbf{e}_{t}^{1:N^2}) \\
    & \quad\quad\quad\quad\quad\,\, \cdot p(\mathbf{c}_{t}^{1:N}|\mathbf{c}_{t-1}^{1:N},\mathbf{z}_{t-1}^{1:N})p(\mathbf{e}_{t}^{1:N^2}|\mathbf{e}_{t-1}^{1:N^2},\mathbf{z}_{t-1}^{1:N},\mathbf{y}_{t-1}^{1:N}) \Bigg] \\
    & = \sum_{\mathbf{z}_{t-1}^{1:N},\mathbf{c}_{t-1}^{1:N}}\Bigg[\underline{\alpha_{t-1}(\mathbf{z}_{t-1},\mathbf{c}_{t-1})} \cdot\prod_{n=1}^{N}p(\mathbf{y}_{t}^{n}|\mathbf{y}_{t-1}^{n},z_{t}^{n}) \cdot \prod_{n=1}^{N}\!\prod_{m=1}^{N}\!p(z_{t}^{n}|z_{t-1}^{m}, c_{t}^{n}, \mathbf{y}_{t-1}^{m}, \mathbf{y}_{t-1}^{n}, e_{t}^{m\rightarrow n}) \cdot  \\
    & \quad\quad\quad\quad\quad\,\,  \cdot \prod_{n=1}^{N}p(c_{t}^{n}|c_{t-1}^{n},z_{t-1}^{n})\prod_{n=1}^{N}\!\prod_{m=1}^{N}\!p(e_{t}^{m\rightarrow n}|e_{t-1}^{m\rightarrow n}, z_{t-1}^{m}, z_{t-1}^{n}, c_{t-1}^{m}, c_{t-1}^{n}, \mathbf{y}_{t-1}^{m}, \mathbf{y}_{t-1}^{n})\Bigg], 
\end{align*}
where $\alpha_t(\mathbf{z}_t,\mathbf{c}_t)$ are calculated by $\alpha_{t-1}(\mathbf{z}_{t-1},\mathbf{c}_{t-1})$ recursively with states transitions.

The derivatives of the backward section $\beta_t(\mathbf{z}_t,\mathbf{c}_{t})$ is calculated as
\begin{align*}
    \beta_T(\mathbf{z}_T,\mathbf{c}_T) &= 1 \\
    \underline{\beta_t(\mathbf{z}_t,\mathbf{c}_t)} &= p(\mathbf{y}_{t+1:T},\mathbf{e}_{t+1:T}|\mathbf{y}_{t},\mathbf{z}_t,\mathbf{c}_t) \\
    &= p(\mathbf{y}_{t+1:T}^{1:N},\mathbf{e}_{t+1:T}^{1:N^2}|\mathbf{y}_{t}^{1:N}, \mathbf{z}_{t}^{1:N}, \mathbf{c}_{t}^{1:N})\\
    &= \sum_{\mathbf{z}_{t+1}^{1:N},\mathbf{c}_{t+1}^{1:N}}p(\mathbf{y}_{t+1:T}^{1:N},\mathbf{e}_{t+1:T}^{1:N^2},\mathbf{z}_{t+1}^{1:N},\mathbf{c}_{t+1}^{1:N} |\mathbf{y}_{t}^{1:N}, \mathbf{z}_{t}^{1:N}, \mathbf{c}_{t}^{1:N})\\
    &= \sum_{\mathbf{z}_{t+1}^{1:N},\mathbf{c}_{t+1}^{1:N}} \Bigg[p(\mathbf{y}_{t+1}^{1:N}|\mathbf{y}_{t}^{1:N},\mathbf{z}_{t+1}^{1:N})p(\mathbf{z}_{t+1}^{1:N}|\mathbf{z}_{t}^{1:N},\mathbf{c}_{t+1}^{1:N},\mathbf{y}_{t}^{1:N},\mathbf{e}_{t+1}^{1:N^2}) \\
    & \cdot p(\mathbf{c}_{t+1}^{1:N}|\mathbf{c}_{t}^{1:N},\mathbf{z}_{t}^{1:N})p(\mathbf{e}_{t+1}^{1:N^2}|\mathbf{e}_{t}^{1:N^2},\mathbf{z}_{t}^{1:N},\mathbf{y}_{t}^{1:N})p(\mathbf{y}_{t+2:T}^{1:N},\mathbf{e}_{t+2:T}^{1:N^2}|\mathbf{y}_{t+1}^{1:N}, \mathbf{z}_{t+1}^{1:N}, \mathbf{c}_{t+1}^{1:N}) \Bigg]\\
    &= \sum_{\mathbf{z}_{t+1}^{1:N},\mathbf{c}_{t+1}^{1:N}}\Bigg[ \prod_{n=1}^{N}p(\mathbf{y}_{t+1}^{n}|\mathbf{y}_{t}^{n},z_{t+1}^{n}) \cdot \prod_{n=1}^{N}\!\prod_{m=1}^{N}\!p(z_{t+1}^{n}|z_{t}^{m}, c_{t+1}^{n}, \mathbf{y}_{t}^{m,n}, e_{t+1}^{m\rightarrow n}) \\
    & \cdot \prod_{n=1}^{N}p(c_{t+1}^{n}|c_{t}^{n},z_{t}^{n}) \cdot \prod_{n=1}^{N}\!\prod_{m=1}^{N}\!p(e_{t+1}^{m\rightarrow n}|e_{t}^{m\rightarrow n}, z_{t}^{m}, z_{t}^{n}, c_{t}^{m}, c_{t}^{n}, \mathbf{y}_{t}^{m},\mathbf{y}_{t}^{n})\,\,\underline{\beta_{t+1}(\mathbf{z}_{t+1},\mathbf{c}_{t+1})}\Bigg],
\end{align*}
where $\beta_t(\mathbf{z}_t,\mathbf{c}_t)$ is computed via $\beta_{t+1}(\mathbf{z}_{t+1},\mathbf{c}_{t+1})$ recursively by state transitions.

\subsection{Derivation of Optimization Objective}
\label{app: Derivation of Optimization Objective of AMORE-MIO}
Learnable parameters of our model are optimized by maximizing the evidence lower bound (ELBO) with sparse regularization on coefficients of candidate basis functions where the derivatives of ELBO are as follows.
For brevity, $\mathbf{y}$, $\mathbf{z}$, $\mathbf{c}$, and $\mathbf{e}$ represents $\mathbf{y}_{1:T}^{1:N}$, $\mathbf{z}_{1:T}^{1:N}$, $\mathbf{c}_{1:T}^{1:N}$, and $\mathbf{e}_{1:T}^{1:N^2}$ respectively. $N$ is the number of objects. $T$ is the number of time steps.
\begin{align*}
ELBO &= {\rm log}\,p_\theta(\mathbf{y})\!-\!D_{K\!L}\left[q_\phi(\mathbf{z},\mathbf{c},\mathbf{e}|\mathbf{y})\,\|\,p_\theta(\mathbf{z},\mathbf{c},\mathbf{e}|\mathbf{y})\right] \\
&= \int q_\phi(\mathbf{z},\mathbf{c},\mathbf{e}|\mathbf{y}) \, {\rm log}\,p_\theta(\mathbf{y}) \, d(\mathbf{z},\mathbf{c},\mathbf{e}) - \int q_\phi(\mathbf{z},\mathbf{c},\mathbf{e}|\mathbf{y}) \, {\rm log}\,\frac{q_\phi(\mathbf{z},\mathbf{c},\mathbf{e}|\mathbf{y})}{p_\theta(\mathbf{z},\mathbf{c},\mathbf{e}|\mathbf{y})} \, d(\mathbf{z},\mathbf{c},\mathbf{e}) \\
&= \int q_\phi(\mathbf{z},\mathbf{c},\mathbf{e}|\mathbf{y})\left[{\rm log}\,p_\theta(\mathbf{z},\mathbf{c},\mathbf{e},\mathbf{y})-{\rm log}\,q_\phi(\mathbf{z},\mathbf{c},\mathbf{e}|\mathbf{y})\right] \, d(\mathbf{z},\mathbf{c},\mathbf{e}) \\
&= \mathbb{E}_{q_\phi(\mathbf{z},\mathbf{c},\mathbf{e}|\mathbf{y})}\left[{\rm log}\,p_\theta(\mathbf{z},\mathbf{c},\mathbf{e},\mathbf{y})-{\rm log}\,q_\phi(\mathbf{z},\mathbf{c},\mathbf{e}|\mathbf{y})\right] \\
&= \mathbb{E}_{q_\phi(\mathbf{e}|\mathbf{y})p_\theta(\mathbf{z},\mathbf{c}|\mathbf{y},\mathbf{e})}\left[{\rm log}\,p_\theta(\mathbf{y},\mathbf{e})p_\theta(\mathbf{z},\mathbf{c}|\mathbf{y},\mathbf{e})-{\rm log}\,q_\phi(\mathbf{e}|\mathbf{y})p_\theta(\mathbf{z},\mathbf{c}|\mathbf{y},\mathbf{e})\right] \\
&= \mathbb{E}_{q_\phi(\mathbf{e}|\mathbf{y})}\left[{\rm log}\,p_\theta(\mathbf{y},\mathbf{e})-{\rm log}\,q_\phi(\mathbf{e}|\mathbf{y})\right] \\
&= 
\mathbb{E}_{q_\phi(\mathbf{e}|\mathbf{y})}\left[{\rm log}\,p_\theta(\mathbf{y},\mathbf{e})\right]+H(q_\phi(\mathbf{e}|\mathbf{y})),
\end{align*}
where ${\rm log}\,p_\theta(\mathbf{y},\mathbf{e})$ is a joint likelihood, and $H(q_\phi(\mathbf{e}|\mathbf{y}))$ is a conditional entropy for the approximate posterior of edge variable $\mathbf{e}$.
\subsubsection{Training of ELBO}
We use the mini-batch stochastic gradient descent algorithm for training of ELBO. The gradients with respect to $\theta$ or $\phi$ in ELBO are calculated as
\begin{align*}
 \nabla_\theta ELBO &= \nabla_\theta \left[\mathbb{E}_{q_\phi(\mathbf{e}|\mathbf{y})}{\rm log}\,p_\theta(\mathbf{y},\mathbf{e})\right] = \mathbb{E}_{q_\phi(\mathbf{e}|\mathbf{y})}\nabla_\theta{\rm log}\,p_\theta(\mathbf{y},\mathbf{e}), \\
\nabla_\phi ELBO &= \nabla_\phi \left[\mathbb{E}_{q_\phi(\mathbf{e}|\mathbf{y})}{\rm log}\,p_\theta(\mathbf{y},\mathbf{e})+H(q_\phi(\mathbf{e}|\mathbf{y}))\right] \\
& = \nabla_\phi \left[\mathbb{E}_{q_\phi(\mathbf{e}|\mathbf{y})}{\rm log}\,p_\theta(\mathbf{y},\mathbf{e})\right] + \nabla_\phi H(q_\phi(\mathbf{e}|\mathbf{y}))\\
& = \mathbb{E}_{\,\mathbf{\epsilon} \sim \mathcal{N}}\left[\nabla_\phi{\rm log}\,p_\theta(\mathbf{e},\mathbf{y}_\phi(\mathbf{e}, \mathbf{\epsilon}))\right] + \nabla_\phi H(q_\phi(\mathbf{e}|\mathbf{y})),
\end{align*}
where we use the reparameterization trick~\cite{kingma2013auto} to calculate the gradients of $\nabla_\phi \left[\mathbb{E}_{q_\phi(\mathbf{e}|\mathbf{y})}{\rm log}\,p_\theta(\mathbf{y},\mathbf{e})\right]$. 
$\nabla_\phi H(q_\phi(\mathbf{e}|\mathbf{y}))$ is an entropy loss.
Among the derivative terms, the challenging part is the gradients of joint probability $\nabla_\theta{\rm log}\,p_\theta(\mathbf{y},\mathbf{e})$, which is calculated as
\begin{align*}
\nabla{\rm log}\,p(\mathbf{y},\mathbf{e}) 
&= \mathbb{E}_{p(\mathbf{z},\mathbf{c}|\mathbf{y}, \mathbf{e})}\left[\nabla{\rm log}\,p(\mathbf{y},\mathbf{e})\right] \\
&= \mathbb{E}_{p(\mathbf{z},\mathbf{c}|\mathbf{y},\mathbf{e})}\left[\nabla{\rm log}\,p(\mathbf{y},\mathbf{e},\mathbf{z},\mathbf{c})\right] - \mathbb{E}_{p(\mathbf{z},\mathbf{c}|\mathbf{y},\mathbf{e})}\left[\nabla{\rm log}\,p(\mathbf{z},\mathbf{c}|\mathbf{y},\mathbf{e})\right] \\
&= \mathbb{E}_{p(\mathbf{z},\mathbf{c}|\mathbf{y},\mathbf{e})}\left[\nabla{\rm log}\,p(\mathbf{y},\mathbf{e},\mathbf{z},\mathbf{c})\right] - \int p(\mathbf{z},\mathbf{c}|\mathbf{y},\mathbf{e})\frac{\nabla{\rm log}\,p(\mathbf{z},\mathbf{c}|\mathbf{y},\mathbf{e})}{p(\mathbf{z},\mathbf{c}|\mathbf{y},\mathbf{e})}d(\mathbf{z},\mathbf{c})\\
&= \mathbb{E}_{p(\mathbf{z},\mathbf{c}|\mathbf{y},\mathbf{e})}\left[\nabla{\rm log}\,p(\mathbf{y},\mathbf{e},\mathbf{z},\mathbf{c})\right],
\end{align*}
Following the Markovian property, we unfold the joint likelihood $p(\mathbf{y},\mathbf{e},\mathbf{z},\mathbf{c})$ over time as:
  \begin{align*}
    &\nabla{\rm log}\,p(\mathbf{y},\mathbf{e},\mathbf{z},\mathbf{c}) \\
    &= \nabla\,{\rm log}\,p(\mathbf{y}_{1:T}^{1:N},\mathbf{e}_{1:T}^{1:N^2},\mathbf{z}_{1:T}^{1:N},\mathbf{c}_{1:T}^{1:N}) \\
    &= \nabla\,{\rm log}\!\left[p(\mathbf{y}_{1}^{1:N}|\mathbf{z}_{1}^{1:N})p(\mathbf{z}_{1}^{1:N})\right] + \sum_{t=2}^{T}\nabla\,{\rm log}\!\bigg[p(\mathbf{y}_{t}^{1:N}|\mathbf{y}_{t-1}^{1:N},\mathbf{z}_{t}^{1:N})p(\mathbf{z}_{t}^{1:N}|\mathbf{z}_{t-1}^{1:N},\mathbf{c}_{t}^{1:N},\mathbf{y}_{t-1}^{1:N},\mathbf{e}_{t}^{1:N^2})\cdot\\ 
    &\;\;\;\;\;\;\;\;p(\mathbf{c}_{t}^{1:N}|\mathbf{c}_{t-1}^{1:N},\mathbf{z}_{t-1}^{1:N})p(\mathbf{e}_{t}^{1:N^2}|\mathbf{e}_{t-1}^{1:N^2},\mathbf{z}_{t-1}^{1:N},\mathbf{y}_{t-1}^{1:N})\bigg] \\
    &= \nabla\,{\rm log}\!\left[\prod_{n=1}^{N}p(\mathbf{y}_{1}^{n}|z_{1}^{n})\cdot \prod_{n=1}^{N}p(z_{1}^n)\right] + \sum_{t=2}^{T}\nabla\,{\rm log}\!\Bigg[\prod_{n=1}^{N}p(\mathbf{y}_{t}^{n}|\mathbf{y}_{t-1}^{n},z_{t}^{n}) \cdot \prod_{n=1}^{N}\!\prod_{m=1}^{N}\!p(z_{t}^{n}|z_{t-1}^{m}, c_{t}^{n}, \mathbf{y}_{t-1}^{m},\mathbf{y}_{t-1}^{n}, e_{t}^{m\rightarrow n}) \cdot \\
    &\;\;\;\;\;\;\;\;\prod_{n=1}^{N}p(c_{t}^{n}|c_{t-1}^{n},z_{t-1}^{n}) \cdot \prod_{n=1}^{N}\!\prod_{m=1}^{N}\!
    p(e_{t}^{m \rightarrow n}|e_{t-1}^{m \rightarrow n}, z_{t-1}^{m}, z_{t-1}^{n}, c_{t-1}^{m}, c_{t-1}^{n}, \mathbf{y}_{t-1}^{m}, \mathbf{y}_{t-1}^{n}) \Bigg],
\end{align*}
where edge variables evolve based on all previous states of both objects. We model the influences of interactions between each pair of objects by $p(z_{t}^{n}|z_{t-1}^{m}, c_{t}^{n}, \mathbf{y}_{t-1}^{m}, \mathbf{y}_{t-1}^{n}, e_t^{m\rightarrow n})$ without instantaneous dependences. Combining with expectation, $\nabla{\rm log}\,p(\mathbf{y},\mathbf{e})$ is finally calculated as
\begin{align*}
    \nabla{\rm log}\,p(\mathbf{y},\mathbf{e})
    &=\mathbb{E}_{p(\mathbf{z},\mathbf{c}|\mathbf{y},\mathbf{e})}\left[\nabla{\rm log}\,p(\mathbf{y},\mathbf{e},\mathbf{z},\mathbf{c})\right]\\
    &= \mathbb{E}_{p(\mathbf{z}_{1:T}^{1:N},\mathbf{c}_{1:T}^{1:N}|\mathbf{y}_{1:T}^{1:N},\mathbf{e}_{1:T}^{1:N^2})}\left[\nabla{\rm log}\,p(\mathbf{y}_{1:T}^{1:N},\mathbf{e}_{1:T}^{1:N^2},\mathbf{z}_{1:T}^{1:N},\mathbf{c}_{1:T}^{1:N})\right] \\
    &= \sum_{\mathbf{k}}p(\mathbf{z}_1^{1:N}=\mathbf{k}|\mathbf{y}_{1:T}^{1:N},\mathbf{e}_{1:T}^{1:N^2}) \nabla\,{\rm log}\left[\prod_{n=1}^{N}p(\mathbf{y}_{1}^{n}|\mathbf{z}_{1}^{n}=k^n)\cdot p(\mathbf{z}_{1}^{1:N}=\mathbf{k})\right] \\
    &\,\,\,\,\,\,\, + \sum_{t=2}^{T}\sum_{\mathbf{k},\mathbf{j},\mathbf{u},\mathbf{v}}\!\! \xi(\mathbf{k},\mathbf{j},\mathbf{u},\mathbf{v})\,\nabla\,{\rm log}\!\Bigg[\prod_{n=1}^{N}\!\prod_{m=1}^{N}\!p(e_{t}^{m\rightarrow n}|e_{t-1}^{m\rightarrow n}, \mathbf{z}_{t}^{m,n}=\mathbf{k}^{m,n}, \mathbf{y}_{t}^{m,n}) \cdot \prod_{n=1}^{N}p(\mathbf{y}_{t}^{n}|\mathbf{y}_{t-1}^{n},z_{t}^{n}=k^n)\\
    &\,\,\,\,\,\,\,\cdot \prod_{n=1}^{N}\!\prod_{m=1}^{N}\!\,p(z_{t}^{n}\!=\!k^n|z_{t-1}^{m}\!=\!j^m, c_{t}^{n}\!=\!v^n, \mathbf{y}_{t-1}^{m,n}, e_{t-1}^{m\rightarrow n})\cdot \prod_{n=1}^{N}p(c_{t}^{n}\!=\!v^n|c_{t-1}^{n}\!=\!u^n,z_{t-1}^{n}\!=\!j^n)\Bigg]\\
    &= \sum_{\mathbf{k}}\gamma(\mathbf{k})\, \nabla\,{\rm log}\!\left[B_1(k^n)\cdot \pi(\mathbf{k})\right] \\
    &\,\,\,\,\,\,\,+\sum_{t=2}^{T}\sum_{\mathbf{k},\mathbf{j},\mathbf{u},\mathbf{v}}\!\!\xi(\mathbf{k},\mathbf{j},\mathbf{u},\mathbf{v})\,\nabla\,{\rm log}\!\left[B_t(\mathbf{k})\cdot 
    E_t(\mathbf{k}) \cdot
    A_t(\mathbf{k},\mathbf{j},\mathbf{v}) \cdot  C_t(\mathbf{u},\mathbf{v},\mathbf{j})\right],
\end{align*}
where $\pi(\mathbf{k})$, $\gamma(\mathbf{k})$, $\xi(\mathbf{k},\mathbf{j},\mathbf{u},\mathbf{v})$, $B_t(\mathbf{k})$, $E_t(\mathbf{k})$, $A_t(\mathbf{k},\mathbf{j},\mathbf{v})$, and $C_t(\mathbf{u},\mathbf{v},\mathbf{j})$ are defined as
\begin{align*}
\pi(\mathbf{k}) &= p(\mathbf{z}_{1}^{1:N}=\mathbf{k}),\\
\gamma(\mathbf{k}) &= p(\mathbf{z}_1^{1:N}=\mathbf{k}|\mathbf{y}_{1:T}^{1:N},\mathbf{e}_{1:T}^{1:N^2}),\\
\xi(\mathbf{k},\mathbf{j},\mathbf{u},\mathbf{v}) &= p(\mathbf{z}_t^{1:N}\!\!=\!\mathbf{k},\mathbf{z}_{t-1}^{1:N}\!\!=\!\mathbf{j},\mathbf{c}_t^{1:N}\!=\!\mathbf{v},\mathbf{c}_{t-1}^{1:N}\!=\!\mathbf{u}|\mathbf{y}_{1:T}^{1:N},\mathbf{e}_{1:T}^{1:N^2}), \\
B_t(\mathbf{k}) &= \prod_{n=1}^{N}p(\mathbf{y}_{t}^{n}|\mathbf{y}_{t-1}^{n},z_{t}^{n}=k^n),\\
E_t(\mathbf{k}) &= \prod_{n=1}^{N}\!\prod_{m=1}^{N}\!p(e_{t}^{m\rightarrow n}|e_{t-1}^{m\rightarrow n}, \mathbf{z}_{t}^{m,n}=\mathbf{k}^{m,n}, \mathbf{y}_{t}^{m,n}),\\
A_t(\mathbf{k},\mathbf{j},\mathbf{v}) &= \prod_{n=1}^{N}\!\prod_{m=1}^{N}\!\,p(z_{t}^{n}\!=\!k^n|z_{t-1}^{m}\!=\!j^m, c_{t}^{n}\!=\!v^n, \mathbf{y}_{t-1}^{m,n}, e_{t-1}^{m\rightarrow n}),\\
C_t(\mathbf{u},\mathbf{v},\mathbf{j}) &= \prod_{n=1}^{N}p(c_{t}^{n}\!=\!v^n|c_{t-1}^{n}\!=\!u^n,z_{t-1}^{n}\!=\!j^n),
\end{align*}
Among these, $\pi(\mathbf{k})$ is the initial joint discrete mode probability. $B_t(\mathbf{k})$ is the observation transition probability conditioned on motion modes $\mathbf{k}$. $E_t(\mathbf{k})$ is the discrete edge transition probability. 
$A_t(\mathbf{k},\mathbf{j},\mathbf{v})$ is the discrete motion mode transition probability. 
$C_t(\mathbf{u},\mathbf{v},\mathbf{j})$ is the mode count transition probability.
Besides, $\gamma(\mathbf{k})$ and $\xi(\mathbf{k},\mathbf{j},\mathbf{u},\mathbf{v})$ are conditional posterior distributions, which can be calculated by the forward-backward algorithm in Appendix~\ref{app: Inference Model of AMORE-MIO}.

\section{More Experiments}
\subsection{Details of Datasets}
\label{app: datasets}
\begin{figure}[htbp]
 \centering
 \setlength{\tabcolsep}{1pt}
 \includegraphics[width=.25\linewidth]{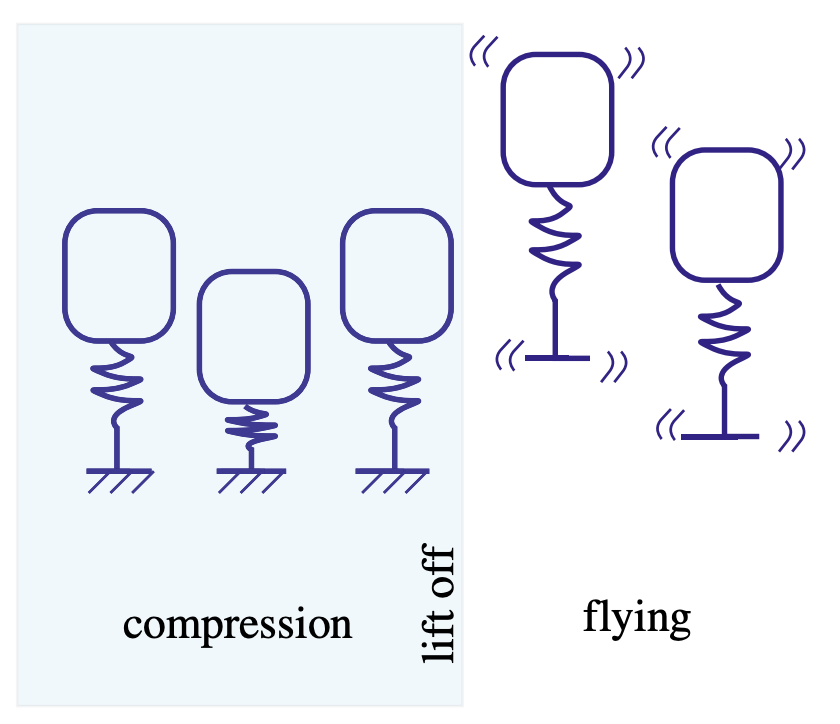}
 \vspace{-8pt}
 \caption{An illustration of Mass-spring hopper system~\citep{brunton2016discovering}.}
 \vspace{-8pt}
 \label{fig: mass-spring-hopping-system}
 \end{figure}
 
\subsubsection{Mass-spring hopper}
\label{app: Mass-spring hopper}
Figure~\ref{fig: mass-spring-hopping-system} shows an illustration of a Mass-spring system that contains two motion modes, i.e. flying and compression. A minimal model of the Mass-spring hopper system is defined as
\begin{align*}
m{\ddot x}
= \begin{cases}
-k(x-x_0) - mg,  &x \leq x_0 \\
-mg,  & x > x_0
\end{cases}\!,
\end{align*}
where $k$, $m$, and $g$ are the spring constant, mass, and gravity, respectively. $x_0$ is the unstretched spring length, which defines the flying $x > x_0$ and compression $x \leq x_0$ modes. After scaling by $\kappa = kx_0/mg$, the equations above becomes
\begin{align*}
{\ddot y}
= \begin{cases}
1 - \kappa(y-1),  &y \leq 1 \\
-1,  & y > 1
\end{cases}\!.
\end{align*}
Following Hybrid-SINDy~\citep{mangan2019model}, we set $\kappa = 10$ for data generation. Denoting $y$ as $l$ and $\dot{y}$ as $v$, thus the target closed-form ordinary differential equations are
\begin{align}
\begin{cases}
\dot{l} = v\;\;\textrm{and}\;\;\dot{v} = 11-10\;l,  &l \leq 1 \\
\dot{l} = v\;\;\textrm{and}\;\;\dot{v} = -1,  & l > 1
\end{cases}
\label{eq: Mass-spring-scaling-final-for-generation}
\end{align}
The generated positions and velocities are concatenated $[l, v]$ and used as observations. Instead of generating only a few samples in Hybrid-SINDy~\citep{mangan2019model} (3 for training and 5 for validation), we scale up the datasets and sample 240 initial conditions from the ranges $(0.5,3)$ and $(-1,1)$ for positions $a$ and velocities $b$, respectively. Among them, 200 samples are for training, 20 for validation, and 20 for testing. The system is simulated to generate 150 time steps for each time series, with sampling intervals of $\triangle_{\tau}=0.033$. We add Gaussian noise with mean zero and standard derivation $10^{-6}$ to generated samples. By default, we use the first 100 time steps as context and predict the following next 50 time steps one by one based on the ground truth of the previous time step, i.e. one-step prediction. By default, the order of polynomial functions is set as 2, and the maximal number of possible modes is 3.

\subsubsection{Susceptible, Infected and Recovered (SIR) Disease Dataset}
\label{app: Susceptible, Infected and Recovered (SIR) Disease Dataset}
The SIR disease model in the epidemiological community has been widely studied in the literature~\cite{toda2020susceptible,mcmahon2020reinfection}. The model can be defined as
\begin{align*}
\dot S &= vN - \frac{\beta_t}{N}IS -dS, \notag \\
\dot I &= \frac{\beta_t}{N}IS - (\gamma + d)I, \notag\\
\dot R &= \gamma I - dR, 
\end{align*}
where the rate of transmission $\beta_t$ is time-varying, which takes two discrete values according to whether the school is in session or not
\begin{align*}
\beta_t
= \begin{cases}
\hat{\beta}\cdot(1+b),  &t \in {\rm school \,\, in \,\, session}, \\
\hat{\beta}/(1+b),  &t \in {\rm school \,\, out \,\, of \,\, session}.
\end{cases}
\end{align*}
Following Hybrid-SINDy~\citep{mangan2019model}, for dataset generation, the rates that define at which students enter and leave the population are set as $v = 1/365$ and $d = 1/365$. The total population of students is set as $N=1000$. The recovery rate is set as $\gamma = 1/5$ assuming 5 days is the average infectious period. The base transmission rate is set as $\hat{\beta}=9.336$ and $b=0.8$ tunes the transmission rate change. 
Following Hybrid-SINDy~\citep{mangan2019model}, the concatenation $[S, I]$ of $S$ and $I$ are used as observations.
Thus the target closed-form ordinary differential equations are
\begin{align}
\begin{cases}
\dot S = 2.74 - 0.0168\;IS -0.0027\;S \;\;{\rm and}\;\; \dot I = 0.0168\;IS - 0.20\;I,  &t \in {\rm school \,\, in \,\, session} \\
\dot S = 2.74 - 0.0052\;IS -0.0027\;S \;\;{\rm and}\;\; \dot I = 0.0052\;IS - 0.20\;I,  &t \in {\rm school \,\, out \,\, of \,\, session}.
\end{cases}
\label{eq: SIR-dataset-exact-number}
\end{align}
In a school year, the in-class periods are 35-155 and 225-365 days. The break periods are 0-35 and 155-225 days. 
Instead of creating only one time series for training and one for validation in Hybrid-SINDy, we scale up the datasets and sample 240 initial conditions for $S_0$, $I_0$, and $R_0$. For instance, in each sample, we first sample a $R_0$ from the range $(900,980)$, and then sample a $I_0$ from the range $(0,1000-R_0)$, and then calculate $S_0$ by $S_0 = 1000 - R_0 - I_0$. We simulate each time series for 2 years with a daily interval, thus producing 730 time steps for each time series. We add a random perturbation to the start of each session by changing the states of $S$, $I$ and $R$ by either -2, -1, 0, 1, or 2, independently. 
By default, we use the first 600 time steps as context and predict the next 130 time steps one by one based on the ground truth of the previous time step, i.e. one-step prediction.
By default, the order of polynomial functions is set as 2, and the maximal number of possible modes is 3 for our methods.

\subsubsection{Non-hybrid Physical Systems}
\label{app: Physical Systems}
Following~\citet{course2023state}, non-hybrid physical systems include the Coupled linear, Cubic oscillator, Lorenz' 63, Hopf bifurcation, Seklov glycolysis, and Duffing oscillator. Equations of a Damped linear oscillator are defined as $\dot{x} = -0.1x + 2y$ and $\dot{y} = -2x - 0.1y$. A Damped cubic oscillator is $\dot{x} = -0.1x^3 + 2y^3$ and $\dot{y} = -2x^3 - 0.1y^3$. A coupled linear system is $\ddot{x} = -6x + 2y$ and $\ddot{y} = 2x - 6y$. A Duffing oscillator is $\dot{x} = y$ and $\dot{y} = -x^3 + x - 0.35y$. A Selkov glycolysis is $\dot{x} = -x + 0.08y + x^2y$ and $\dot{y} = 0.6 - 0.08y - x^2y$. A Lorenz'63 system is $\dot{x} = 10y - 10x$, $\dot{y} = 28x -xz - y$, and $\dot{z} = xy - 2.67z$. A Hopf bifurcation is $\dot{x} = 0.5x + y - x^3 - xy^2$ and $\dot{y} = -x + 0.5y - x^2y - y^3$. We refer readers to see the details in \citep{course2023state}. By default, the order of polynomial functions of the Coupled linear, Cubic oscillator, Lorenz' 63, Hopf bifurcation, Seklov glycolysis, and Duffing oscillator are 2, 3, 2, 3, 3, and 3, respectively for our methods.


\subsubsection{ODE-driven Particle Dataset}
\label{app: ODE-driven Particle Dataset}
Following GRASS~\citep{liu2023graph}, Ordinary Differential Equations are introduced as motion modes to generate trajectories of particles, i.e. Lotka-Volterra, Spiral, and Bouncing Ball
\begin{align}
    &{\rm Lotka\!-\!Volterra\!:}\,\,\dot{x} = x - xy;\,\,\dot{y} = -y + xy, \notag\\
    &{\rm Spiral\!:}\,\,\dot{x} = -0.1x^3 + 2y^3;\,\,\dot{y} = -2x^3 - 0.1y^3, \notag\\
    &{\rm Bouncing\,\,Ball+:}\,\,\dot{x} = 0;\,\,\dot{y} = 2, \notag\\
    &{\rm Bouncing\,\,Ball-:}\,\,\dot{x} = 0;\,\,\dot{y} = -2
\label{eq: ODE-driven Particle Dataset}
\end{align}
Balls are introduced on a squared 2d canvas of size $64*64$ which are with radius $r$ and whose locations are randomly initialized.
Trajectories of balls are generated by numerical values of different equations over time which are mapped to the canvas field.
To simulate mode-switching behaviors, the driven ODE modes of two objects are switched when they collide in the canvas. Different from GRASS~\citep{liu2023graph}, one mode ${\rm Bouncing\,\,Ball}$ is regarded as two modes ${\rm Bouncing\,\,Ball+}$ and ${\rm Bouncing\,\,Ball-}$ in this work as they have different explicit equations for equation discovery.
In summary, 4,928 samples are for training, 191 samples for validation, and 204 samples for testing. Each trajectory has 150 time steps with 10 frames per second. 
By default, the order of polynomial functions is set as 3, and the maximal number of possible modes is 5 for our methods.

\subsubsection{Salsa-dancing Dataset}
\label{app: Salsa-dancing Dataset}
Following GRASS~\citep{liu2023graph}, four modes are annotated and used in the Salsa-dancing dataset, i.e. ``moving forward'', ``moving backward'', ``clockwise turning'', and ``counter-clockwise turning''. In summary, 1,321 samples are for training and 156 samples are for testing. Each sample has 100 time steps, among which 80 for context and the remaining 20 for prediction with 5 frames per second. The coordinates of the skeletal joints of dancers in 3D space are as observations. In practice, for all methods, we utilize two representative joints, i.e. right hip and left hip. By default, the order of polynomial functions is set as 3, and the maximal number of possible modes is 5 for our methods.

\subsection{More Implementation Details}
\label{app: More Implementation Details}

For each dataset, we set different numbers of modes $K$ and orders of polynomial functions $D$ for our model. By default, $K = 3$ and $D = 2$ for the Mass-spring Hopper dataset. $K = 3$ and $D = 2$ for the SIR dataset. $D$ of the Coupled linear, Cubic oscillator, Lorenz’ 63, Hopf bifurcation, Seklov glycolysis, and Duffing oscillator are 2, 3, 2, 3, 3, and 3, respectively. $K = 5$ and $D = 3$ for the ODE-driven particle dataset. $K = 5$ and $D = 3$ for the Salsa-dancing dataset.

\subsection{Statistics of Experiments}
\subsubsection{Mass-spring Hopper}
Experiments with statistics on the Mass-spring Hopper dataset are reported in Tables~\ref{Mass-spring Hopping Dataset segmentation with statistics table} and~\ref{Mass-spring Hopping Dataset prediction with statistics table}, which are extended versions of Tables~\ref{Mass-spring Hopping Dataset segmentation table} and~\ref{Mass-spring Hopping Dataset prediction table} in the main paper.

\begin{table}[ht]
	\centering
        \footnotesize
	\tabcolsep=0.100cm
	\setlength\arrayrulewidth{1.0pt}
	\caption{Segmentation results with statistics on Mass-spring Hopper dataset.}
	\vspace{4pt}
	\begin{tabular}{lcccc}
		\toprule
     Method & NMI $\uparrow$ & ARI $\uparrow$ & Accuracy $\uparrow$ & $F_1$ $\uparrow$ \\
		\midrule
        Hybrid-SINDy  & 0.426 & 0.383 & 0.705 & 0.691 \cr
        AMORE (ours) &  \textbf{0.928$\pm$0.011} & \textbf{0.967$\pm$0.013} & \textbf{0.991$\pm$0.005} & \textbf{0.993$\pm$0.007} \cr
		\bottomrule
	\end{tabular}
	\label{Mass-spring Hopping Dataset segmentation with statistics table}
\end{table}

\begin{table}[ht]
	\centering
        \footnotesize
	\tabcolsep=0.200cm
	\setlength\arrayrulewidth{1.0pt}
	\caption{Forecasting results with statistics on Mass-spring Hopper dataset.}
	\vspace{4pt}
	\begin{tabular}{lcc}
		\toprule
     Method & NMAE $\downarrow$ & NRMSE $\downarrow$ \\
		\midrule
        LLMTime & 0.113$\pm$0.032 / 0.305$\pm$0.036 & 0.417$\pm$0.051 / 0.454$\pm$0.072 \cr
        SVI  & 0.068$\pm$0.016 / 0.075$\pm$0.011 &  0.148$\pm$0.023 / 0.262$\pm$0.030  \cr
        AMORE (ours) & \textbf{0.008$\pm$0.003} / \textbf{0.039$\pm$0.008} & \textbf{0.026$\pm$0.005} / \textbf{0.059$\pm$0.006}  \cr
		\bottomrule
	\end{tabular}
	\label{Mass-spring Hopping Dataset prediction with statistics table}
\end{table}

\subsubsection{SIR disease}
Experiments with statistics on the SIR disease dataset are reported in Tables~\ref{SIR Disease Dataset segmentation with statistics table} and~\ref{SIR Disease Dataset prediction with statistics table}, which are extended versions of Tables~\ref{SIR Disease Dataset segmentation table} and~\ref{SIR Disease Dataset prediction table} in the main paper.

\begin{table}[ht]
	\centering
        \footnotesize
	\tabcolsep=0.100cm
	\setlength\arrayrulewidth{1.0pt}
	\caption{Segmentation results with statistics on the SIR disease dataset.}
	\vspace{4pt}
	\begin{tabular}{lcccc}
		\toprule
     Method & NMI $\uparrow$ & ARI $\uparrow$ & Accuracy $\uparrow$ & $F_1$ $\uparrow$ \\
		\midrule
        Hybrid-SINDy  & 0.296 & 0.283 & 0.538 & 0.519 \cr
        AMORE (ours) &  \textbf{0.475$\pm$0.027} & \textbf{0.483$\pm$0.032} & \textbf{0.731$\pm$0.054} & \textbf{0.735$\pm$0.051} \cr
		\bottomrule
	\end{tabular}
	\label{SIR Disease Dataset segmentation with statistics table}
\end{table}

\begin{table}[ht]
	\centering
        \footnotesize
	\tabcolsep=0.200cm
	\setlength\arrayrulewidth{1.0pt}
	\caption{Forecasting results of Susceptible/Infected with statistics on the SIR disease dataset.}
	\vspace{4pt}
	\begin{tabular}{lcc}
		\toprule
     Method & NMAE $\downarrow$ & NRMSE $\downarrow$ \\
		\midrule
        LLMTime & 0.352$\pm$0.073 / 0.396$\pm$0.091 & 0.481$\pm$0.084 / 0.523$\pm$0.096 \cr
        SVI  & 0.257$\pm$0.031 / 0.273$\pm$0.054 &  0.355$\pm$0.050 / 0.401$\pm$0.078  \cr
        AMORE (ours) & \textbf{0.088$\pm$0.012} / \textbf{0.113$\pm$0.018} & \textbf{0.142$\pm$0.029} / \textbf{0.181$\pm$0.035}  \cr
		\bottomrule
	\end{tabular}
	\label{SIR Disease Dataset prediction with statistics table}
\end{table}

\subsection{Additional Ablation Studies}


\subsubsection{Sampling intervals Analysis}
In our experiments, we followed the experimental setup of Hybrid-SINDy on the sampling intervals of the Mass-spring Hopper dataset and the SIR disease dataset. That means we use their standard sampling intervals $\Delta_t$, e.g. $\Delta_t = 0.033$ on the Mass-spring Hopper dataset.  In Table~\ref{Analyses of delta_t on segmentation results of the Mass-spring Hopper dataset table}, we report the segmentation comparison results when $\Delta_t$ increases. We double the previous $\Delta_t$ each time and thus get $\{0.033,0.066,0.132,0.264\}$.
We can see that when $\Delta_t \geq 0.132$, the segmentation performance of Hybrid-SINDy decreases considerably due to the temporal pattern disruption, while our model has a smaller decrease in performance. 
When $\Delta_t$ increases (e.g. $\Delta_t \geq 0.132$), the discretization obviously disrupts the original temporal patterns of time series. Thus, after learning on the discretization, the model shows significantly decreased performance on \emph{labels that are annotated based on the original temporal patterns}. 

\begin{table}[H]
	\centering
        \footnotesize
	\tabcolsep=0.100cm
	\setlength\arrayrulewidth{1.0pt}
	\caption{Analyses of $\Delta_t$ on segmentation results of the Mass-spring Hopper dataset.}
	\vspace{4pt}
	\begin{tabular}{lccccc}
		\toprule
     Sampling interval $\Delta_t$ & Method & NMI $\uparrow$ & ARI $\uparrow$ & Accuracy $\uparrow$ & $F_1$ $\uparrow$ \\
		\midrule
       0.033 & Hybrid-SINDy  & 0.426 & 0.383 & 0.705 & 0.691 \cr
       0.033 & AMORE (ours) &  \textbf{0.928$\pm$0.011} & \textbf{0.967$\pm$0.013} & \textbf{0.991$\pm$0.005} & \textbf{0.993$\pm$0.007} \cr
       \midrule
       0.066 & Hybrid-SINDy  & 0.422 & 0.385 & 0.701 & 0.697 \cr
       0.066 & AMORE (ours) &  \textbf{0.925$\pm$0.017} & \textbf{0.973$\pm$0.014} & \textbf{0.986$\pm$0.007} & \textbf{0.982$\pm$0.010} \cr
       \midrule
       0.132 & Hybrid-SINDy  & 0.235 & 0.201 & 0.447 & 0.413 \cr
       0.132 & AMORE (ours) &  \textbf{0.458$\pm$0.021} & 
       \textbf{0.369$\pm$0.016} & \textbf{0.627$\pm$0.013} & \textbf{0.644$\pm$0.017} \cr
       \midrule
       0.264 & Hybrid-SINDy  & 0.226 & 0.183 & 0.382 & 0.376 \cr
       0.264 & AMORE (ours) &  \textbf{0.417$\pm$0.015} & 
       \textbf{0.335$\pm$0.008} & \textbf{0.574$\pm$0.020} & \textbf{0.580$\pm$0.012} \cr
		\bottomrule
	\end{tabular}
	\label{Analyses of delta_t on segmentation results of the Mass-spring Hopper dataset table}
\end{table}


\subsubsection{ Number of Training Samples Analysis}
To answer the question: ``Given the significantly smaller datasets used by Hybrid-SINDy, can the proposed method maintain this level of performance difference?'', we rerun experiments on the Mass-spring Hopper dataset by varying the number of samples in the training set from 3 (the same as Hybrid-SINDy) to 20 and 200. The comparison results are summarized in Tables~\ref{Segmentation results on the Mass-spring Hopper dataset with varying numbers of training samples table},~\ref{Forecasting results of Location/Velocity on the Mass-spring Hopper dataset with varying numbers of training samples table}, and~\ref{Reconstruction errors (RER) of discovered equations on the Mass-spring Hopper dataset with varying numbers of training samples table}. In the few-shot setting with a very low number of samples, e.g. 3 samples, Hybrid-SINDy outperforms AMORE. This is expected and a common limitation of deep learning methods, which usually require larger numbers of samples for training. On the other hand, when given more samples, e.g. larger than 20, AMORE outperforms Hybrid-SINDy consistently. 

\begin{table}[H]
	\centering
        \footnotesize
	\tabcolsep=0.100cm
	\setlength\arrayrulewidth{1.0pt}
	\caption{Analyses of numbers of training samples on segmentation results of the Mass-spring Hopper dataset.}
	\vspace{4pt}
	\begin{tabular}{lccccc}
		\toprule
     Number of training samples & Method & NMI $\uparrow$ & ARI $\uparrow$ & Accuracy $\uparrow$ & $F_1$ $\uparrow$ \\
		\midrule
       3 & Hybrid-SINDy  & \textbf{0.425} & \textbf{0.377} & \textbf{0.693} & \textbf{0.684} \cr
       3 & AMORE (ours) &  0.238$\pm$0.052 & 0.217$\pm$0.065 & 0.474$\pm$0.134 & 0.429$\pm$0.110 \cr
       \midrule
       20 & Hybrid-SINDy  & 0.422 & 0.383 & 0.698 & 0.693 \cr
       20 & AMORE (ours) &  \textbf{0.774$\pm$0.037} & \textbf{0.762$\pm$0.025} & \textbf{0.846$\pm$0.094} & \textbf{0.853$\pm$0.071} \cr
       \midrule
       200 & Hybrid-SINDy  & 0.426 & 0.383 & 0.705 & 0.691 \cr
       200 & AMORE (ours) &  \textbf{0.928$\pm$0.011} & 
       \textbf{0.967$\pm$0.013} & \textbf{0.991$\pm$0.005} & \textbf{0.993$\pm$0.007} \cr
		\bottomrule
	\end{tabular}
	\label{Segmentation results on the Mass-spring Hopper dataset with varying numbers of training samples table}
\end{table}

\begin{table}[H]
	\centering
        \footnotesize
	\tabcolsep=0.100cm
	\setlength\arrayrulewidth{1.0pt}
	\caption{Analyses of numbers of training samples on forecasting results of Location/Velocity on the Mass-spring Hopper dataset.}
	\vspace{4pt}
	\begin{tabular}{lccc}
		\toprule
     Number of training samples & Method & NMAE $\downarrow$  & NRMSE $\downarrow$ \\
		\midrule
       3 & LLMTime  & 0.113$\pm$0.032 / 0.305$\pm$0.036 & 0.417$\pm$0.051 / 0.454$\pm$0.072 \cr
       3 & SVI  & 0.173$\pm$0.039 / 0.341$\pm$0.053 & 0.450$\pm$0.081 / 0.481$\pm$0.094  \cr
       3 & AMORE (ours)  &  \textbf{0.091$\pm$0.018} / \textbf{0.160$\pm$0.026} & \textbf{0.315$\pm$0.049} / \textbf{0.348$\pm$0.042} \cr
       \midrule
       20 & LLMTime  & 0.113$\pm$0.032 / 0.305$\pm$0.036 & 0.417$\pm$0.051 / 0.454$\pm$0.072 \cr
       20 & SVI  & 0.094$\pm$0.020 / 0.147$\pm$0.024 & 0.302$\pm$0.038 / 0.381$\pm$0.044  \cr
       20 & AMORE (ours)  &  \textbf{0.036$\pm$0.012} / \textbf{0.057$\pm$0.018} & \textbf{0.106$\pm$0.025} / \textbf{0.129$\pm$0.031} \cr
       \midrule
       200 & LLMTime  & 0.113$\pm$0.032 / 0.305$\pm$0.036 & 0.417$\pm$0.051 / 0.454$\pm$0.072 \cr
       200 & SVI  & 0.068$\pm$0.016 / 0.075$\pm$0.011 & 0.148$\pm$0.023 / 0.262$\pm$0.030  \cr
       200 & AMORE (ours)  &  \textbf{0.008$\pm$0.003} / \textbf{0.039$\pm$0.008} & \textbf{0.026$\pm$0.005} / \textbf{0.059$\pm$0.006} \cr
       \bottomrule
	\end{tabular}
	\label{Forecasting results of Location/Velocity on the Mass-spring Hopper dataset with varying numbers of training samples table}
\end{table}

\begin{table}[H]
	\centering
        \footnotesize
	\tabcolsep=0.100cm
	\setlength\arrayrulewidth{1.0pt}
	\caption{Analyses of numbers of training samples on reconstruction errors (RER) of discovered equations on the Mass-spring Hopper dataset. Numbers are of $e^{-3}$.}
	\vspace{4pt}
	\begin{tabular}{lcc}
		\toprule
     Number of training samples & Method & RER ($e^{-3}$)  $\downarrow$ \\
		\midrule
       3 & Hybrid-SINDy  & \textbf{8.3} \cr
       3 & AMORE (ours)  &  $17.2\pm2.4$ \cr
       \midrule
       20 & Hybrid-SINDy  & 8.2 \cr
       20 & AMORE (ours)  &  \textbf{5.1$\pm$0.6} \cr
       \midrule
       200 & Hybrid-SINDy  & 7.5 \cr
       200 & AMORE (ours)  &  \textbf{2.4$\pm$0.3} \cr
       \bottomrule
	\end{tabular}
	\label{Reconstruction errors (RER) of discovered equations on the Mass-spring Hopper dataset with varying numbers of training samples table}
\end{table}

\subsubsection{Count Variables Analysis}
\label{app: Count Variables Analysis}
The count variables are introduced by REDSDS~\cite{ansari2021deep} to learn the duration distributions of each mode from the data and to avoid frequent mode-switching behaviors. We show below some ablations studies on count variables in the Mass-spring Hopper system, where the flying mode usually takes more than twice as many time steps as the compression mode. 
To quantitatively compare the discovered equations, we first report the equation reconstruction error (RER) for Hybrid-SINDy, AMORE, and AMORE w/o count variable, which are respectively $7.5e^{-3}$, $2.4e^{-4}$, and $2.8e^{-4}$. We can see that with count variables, AMORE has a lower equation reconstruction error than its counterpart without count variables. In Tables~\ref{Updated Mass-spring Hopping Dataset segmentation table} and~\ref{Updated Mass-spring Hopping Dataset prediction table}, we can see that count variables help AMORE learn fewer false-positive mode-switching behaviors, benefitting segmentation and forecasting.

\begin{table}[H]
	\centering
        \footnotesize
	\tabcolsep=0.100cm
	\setlength\arrayrulewidth{1.0pt}
	\caption{Analyse of count variables on segmentation results of the Mass-spring Hopper dataset.}
	\vspace{4pt}
	\begin{tabular}{lcccc}
		\toprule
     Method & NMI $\uparrow$ & ARI $\uparrow$ & Accuracy $\uparrow$ & $F_1$ $\uparrow$ \\
		\midrule
        Hybrid-SINDy  & 0.426 & 0.383 & 0.705 & 0.691 \cr
        AMORE (ours) &  \textbf{0.928$\pm$0.011} & \textbf{0.967$\pm$0.013} & \textbf{0.991$\pm$0.005} & \textbf{0.993$\pm$0.007} \cr
        AMORE w/o count (ours) &  0.903$\pm$0.017 & 0.929$\pm$0.019 & 0.970$\pm$0.012 & 0.975$\pm$0.013 \cr
		\bottomrule
	\end{tabular}
	\label{Updated Mass-spring Hopping Dataset segmentation table}
\end{table}

\begin{table}[H]
	\centering
        \footnotesize
	\tabcolsep=0.200cm
	\setlength\arrayrulewidth{1.0pt}
	\caption{Analyse of count variables on forecasting results of Location/Velocity on the Mass-spring Hopper dataset.}
	\vspace{4pt}
	\begin{tabular}{lcc}
		\toprule
     Method & NMAE $\downarrow$ & NRMSE $\downarrow$ \\
		\midrule
        LLMTime & 0.113$\pm$0.032 / 0.305$\pm$0.036 & 0.417$\pm$0.051 / 0.454$\pm$0.072 \cr
        SVI  & 0.068$\pm$0.016 / 0.075$\pm$0.011 &  0.148$\pm$0.023 / 0.262$\pm$0.030  \cr
        AMORE (ours) & \textbf{0.008$\pm$0.003} / \textbf{0.039$\pm$0.008} & \textbf{0.026$\pm$0.005} / \textbf{0.059$\pm$0.006}  \cr
        AMORE w/o count (ours) & 0.014$\pm$0.004 / 0.046$\pm$0.007 & 0.052$\pm$0.011 / 0.068$\pm$0.014  \cr
		\bottomrule
	\end{tabular}
	\label{Updated Mass-spring Hopping Dataset prediction table}
\end{table}

\subsubsection{Polynomial orders and mode numbers analysis}
To qualitatively show the discovered equations when the order of candidates and the number of modes are increased, we increase the order of candidates from 2 to 5, i.e. $D=5$, and the number of modes from 3 to 5, i.e. $K=5$ on the Mass-spring Hopper dataset. The discovered equations are summarized in Table~\ref{Analyse of equation discovery of AMORE when increasing the number of modes K and the order of candidate basis functions D on the Mass-spring Hopper dataset table}. We can see that our model can categorize exactly 2 modes, i.e. the same as the ground truth, no matter how many potential modes are introduced. Besides, the discovered equations of the 2 modes are regularized by sparsity promotion and do not involve irrelevant function terms thanks to the sparsity regularization when increasing the order of polynomial basis functions.

\begin{table}[H]
	\centering
        \footnotesize
	\tabcolsep=0.200cm
	\setlength\arrayrulewidth{1.0pt}
	\caption{Analyse of equation discovery of AMORE when increasing the number of modes $K$ and the order of candidate basis functions $D$ on the Mass-spring Hopper dataset.}
	\vspace{4pt}
	\begin{tabular}{lcc}
		\toprule
     Settings & Discovered Equations & Ground-truth Equations \\
		\midrule
        $K=3$ and $D=2$ & $\dot{l} = v$ and $\dot{v} = 11.03-10.08l$; $\dot{l} = v$ and $\dot{v} = -1$ & $\dot{l} = v$ and $\dot{v} = 11-10l$; $\dot{l} = v$ and $\dot{v} = -1$ \cr
        $K=5$ and $D=5$ & $\dot{l} = v$ and $\dot{v} = 10.95-10.06l$; $\dot{l} = v$ and $\dot{v} = -1$ & $\dot{l} = v$ and $\dot{v} = 11-10l$; $\dot{l} = v$ and $\dot{v} = -1$ \cr
		\bottomrule
	\end{tabular}
	\label{Analyse of equation discovery of AMORE when increasing the number of modes K and the order of candidate basis functions D on the Mass-spring Hopper dataset table}
\end{table}

\end{document}